\pdfoutput=1
\documentclass{article}
\newcommand{\model}{MetaGS} % model name
\newcommand{\figref}[1]{Figure~\ref{#1}} % 图片引用
\newcommand{\tabref}[1]{Table~\ref{#1}} %表格引用

\usepackage{algorithm}  
\usepackage{algorithmicx}  
\usepackage{algpseudocode}  
\usepackage{multirow}
\usepackage{adjustbox}
\usepackage{comment}
\usepackage{stfloats}
\usepackage{amsmath}
\usepackage{enumitem}
\usepackage{tabularray}
\usepackage{makecell}
\usepackage{ragged2e}
\usepackage{graphicx}
\usepackage{booktabs}
\usepackage{wrapfig}
\usepackage[preprint, nonatbib]{neurips_2025}

\usepackage[utf8]{inputenc} % allow utf-8 input
\usepackage[T1]{fontenc}    % use 8-bit T1 fonts
\usepackage{hyperref}       % hyperlinks
\usepackage{url}            % simple URL typesetting
\usepackage{booktabs}       % professional-quality tables
\usepackage{amsfonts}       % blackboard math symbols
\usepackage{nicefrac}       % compact symbols for 1/2, etc.
\usepackage{microtype}      % microtypography
\usepackage{xcolor}         % colors

\title{MetaGS: A Meta-Learned Gaussian-Phong Model for Out-of-Distribution 3D Scene Relighting}

\author{
Yumeng~He%\footnotemark[1] 
\qquad
Yunbo~Wang\thanks{Corresponding author. 
}
\qquad
Xiaokang~Yang\\
MoE Key Lab of Artificial Intelligence, AI Institute, \\Shanghai Jiao Tong University\\
{\tt \{ymhe, yunbow, xkyang\}@sjtu.edu.cn}
}

\begin{document}

\maketitle

% \section{Submission of papers to NeurIPS 2025}
\begin{abstract}

Out-of-distribution (OOD) 3D relighting requires novel view synthesis under unseen lighting conditions that differ significantly from the observed images. Existing relighting methods, which assume consistent light source distributions between training and testing, often degrade in OOD scenarios. We introduce \textbf{MetaGS} to tackle this challenge from two perspectives. First, we propose a meta-learning approach to train 3D Gaussian splatting, which explicitly promotes learning generalizable Gaussian geometries and appearance attributes across diverse lighting conditions, even with biased training data. Second, we embed fundamental physical priors from the \textit{Blinn-Phong} reflection model into Gaussian splatting, which enhances the decoupling of shading components and leads to more accurate 3D scene reconstruction. Results on both synthetic and real-world datasets demonstrate the effectiveness of MetaGS in challenging OOD relighting tasks, supporting efficient point-light relighting and generalizing well to unseen environment lighting maps.

%in terms of training efficiency and rendering quality 

\end{abstract}
\section{Introduction}

3D scene relighting generates novel lighting effects that interact with the observed 3D environment, with recent advances in learning-based volume rendering offering effective solutions~\cite{Srinivasan_2021_CVPR,Zhang_2021_CVPR,Toschi_2023_CVPR,Zheng_2023_CVPR,Wang_2023_ICCV,sun2023sol,Chang_2024_WACV,Yang_2023_CVPR}.
%Recent advancements in learning-based volume rendering have introduced effective 3D relighting approaches~\cite{Srinivasan_2021_CVPR,Zhang_2021_CVPR,Toschi_2023_CVPR,Zheng_2023_CVPR,Wang_2023_ICCV,sun2023sol,Chang_2024_WACV,Yang_2023_CVPR}.
%
A typical approach captures a substantial number of multi-view images of a scene under individual lighting conditions, then trains a NeRF or 3D Gaussian splatting (3DGS) model to generalize to new point light positions~\cite{wu2024deferredgs, liang2023envidr, jiang2023gaussianshader, zhu2024gs, gao2023relightable}.
Notably, such idealized data requirements are often impractical in real-world scenarios.
To simulate a more realistic capturing process, we build on prior work using a \textbf{One Light At a Time (OLAT)} setup~\cite{ zeng2023relighting,liang2024gs, bi2024gs}, where training data is collected with a moving point light and a moving camera.
Recent studies have further explored low-cost OLAT capture using a smartphone flashlight as the moving light source~\cite{cheng2023wildlight}. 

% However, the OLAT setup accentuates the potential variations in light sources during the practical data collection process, while posing a significant challenge as it increases the difficulty of learning the true object geometries due to the limited training corpus, with entangled time-varying lighting and viewing directions.

% While most OLAT methods assume a coherent lighting distribution between training and testing, real-world scenarios often involve light sources that are randomly distributed around the scene, leading to an \textbf{out-of-distribution (OOD)} issue where test-time light sources deviate from the training distribution. 

\begin{figure}
  \centering
  \includegraphics[width=\columnwidth]{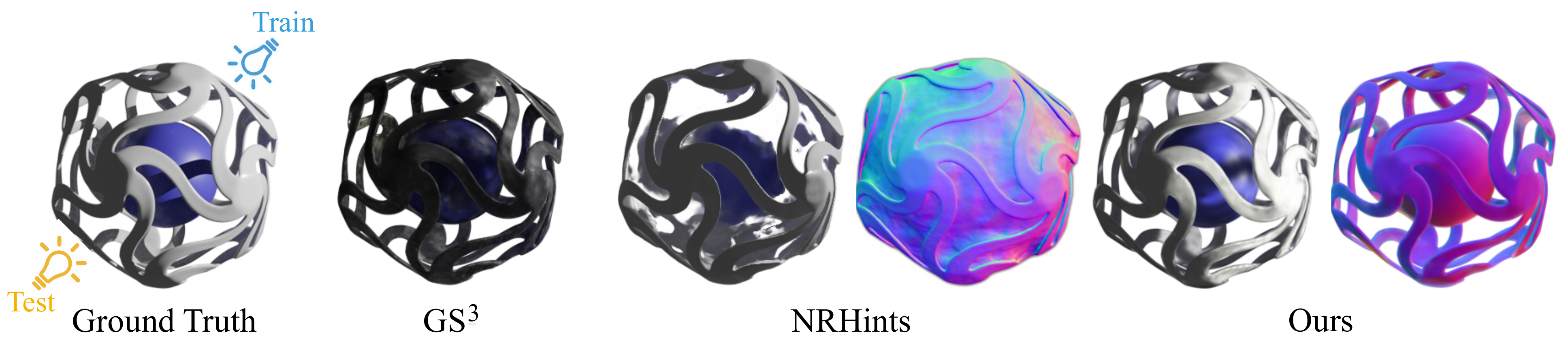}
  \vspace{-15pt}
  \caption{
  \textit{Preliminary results:} NRHints~\cite{zeng2023relighting} and GS$^3$~\cite{bi2024gs} struggle with out-of-distribution light positions (as specified in Section. \ref{sec:preliminary}), as the blurry novel view synthesis results and erroneous normal predictions indicate.
  GS$^3$~\cite{bi2024gs} does not require well-defined surface normals and does not directly output the normal image.
  % Empirical results by evaluating the relighting models with out-of-distribution light positions. We show the synthesized novel view images and normal predictions from NRHints~\cite{zeng2023relighting} and GS-Phong.
  % The results of existing 3DGS-based and NeRF-based relighting techniques (\textit{i.e.} NRHints~\cite{zeng2023relighting}) evaluated on out-of-distribution light positions. (a) ground truth image, (b)-(c) NVS rendering and normal prediction by \model{}, (d) OOD experiment setting, (e)-(f) NVS rendering and normal prediction by \citet{zeng2023relighting}.
  }
  \vspace{-5pt}
  \label{sections/fig:nrhints}
\end{figure}

While most OLAT methods assume a coherent lighting distribution between training and testing, real-world scenarios often involve light sources that are randomly distributed around the scene, which may result in biased training data.
This leads to an \textbf{out-of-distribution (OOD)} issue where test-time light sources deviate from the training distribution. 
This OOD relighting presents a more natural and challenging task, serving as a robust test for evaluating a model's performance under ``\textit{truly}'' novel lighting conditions. 
Compared with the standard OLAT setup, it complicates the learning of true object illumination properties and geometries due to a limited training corpus with entangled, time-varying lighting and viewing directions and, more critically, spatially biased lighting.
% to a limited training corpus with entangled, time-varying lighting and viewing directions plus spatially biased lighting distribution.
% An empirical observation is that the existing methods struggle with OOD relighting. In such cases, as presented in Figure \ref{sections/fig:nrhints}, they often produce unrealistic lighting effects, such as mixing specular highlights with shadows, due to overfitting the training samples.
As shown in Figure \ref{sections/fig:nrhints}, existing OLAT methods struggle with the OOD relighting task, often producing unrealistic lighting effects, such as chaotic specular highlights and shadows, due to overfitting to the training samples.
Also, because of the lack of generalizability, when new lighting distributions are introduced, they typically require full retraining or finetuning to adapt.

% \begin{figure}[t]
%     \centering
%     \includegraphics[width=\linewidth]{sections/fig/olat.pdf}
%     \caption{
%    \textit{The One Light At a Time (OLAT) relighting setup:} The training data is collected using a moving point light and a moving camera. Our model decomposes lighting effects in OLAT and facilitates 3D rendering with out-of-distribution illuminations.
%    % Unlike existing OLAT methods, the proposed \textit{Gaussian-Phong} model incorporates the Phong reflection model into the Gaussian splatting approach. This integration enables the decomposition of various lighting effects (\textbf{right}) and facilitates 3D scene rendering under out-of-distribution lighting conditions.
%    }
%     \label{sections/fig:olat}
% \end{figure}

% In this paper, we present two novel techniques to enhance the generalizability of the OLAT model.

% First, we introduce a simple yet general physical assumption by integrating the classical Blinn-Phong reflection model~\cite{blinn1977models} into the differentiable rendering framework of 3DGS.  This method offers a natural advantage by allowing the model to decouple various shading components of the scene (\figref{sections/fig:olat}). The proposed \textbf{\textit{Gaussian-Phong}} model (\model{}) leads to a more physically grounded understanding of the interactions between objects and lighting in the training data. It aligns with prior research demonstrating that learning modular and compositional representations enhances generalizability.

% Second, we further alleviate the reliance on extensive training data in OLAT by leveraging a \textbf{\textit{meta-learning}} approach to train the relighting model. We frame OLAT as a multi-task learning problem, viewing the rendering under specific lighting positions as separate tasks. The core idea is that the gradient updates of the learned Gaussian geometries under a given lighting condition should be validated using data from different lighting conditions. To achieve this, we employ a bilevel optimization approach that facilitates this gradient validation-and-update process.

% We validate \model{} through experiments on both synthetic and real-world datasets, showcasing that the proposed GS-Phong model and the meta-learning approach jointly enhance generalization to OOD light sources. 

In this paper, we present two novel techniques to enable better OOD relighting.
First, we provide a pilot study on \textbf{meta-learning-based 3DGS} and introduce \model{} to mitigate overfitting and improve the generalizability of the OLAT models.
This approach leverages bilevel optimization techniques to effectively address the challenges associated with limited training data and complex lighting variations. 
Specifically, we frame OLAT as a multi-task learning problem, treating rendering under specific lighting positions as distinct tasks. 
The core idea is to validate the learned Gaussian illumination properties under a given lighting condition using data from other lighting conditions.
The resulting second-order gradients of 3DGS parameters can explicitly encourage the model to generalize to unseen illumination rather than overfitting to specific training samples, as is common with standard 3DGS loss functions.
%
% Notebly, our method does not require extensive multi-light, multi-camera data collection~\cite{bi2024gs,zhang2024prtgaussian}, making it more practical for real-world relighting scenarios.

Furthermore, \model{} integrates simple yet fundamental physical priors into Gaussian splatting by incorporating \textbf{a learnable Blinn-Phong reflection model}~\cite{blinn1977models}. 
This approach effectively decouples different shading components---diffuse, specular, and ambient---leading to a more physically grounded understanding of the interactions between objects and lighting in the training data.
This design draws inspiration from prior research~\cite{zhang2021nerfactor, wu2024deferredgs, liang2023envidr} indicating that modular and compositional representations enhance generalization capabilities.
The decoupled rendering improves the model's ability to generalize to novel lighting conditions, allowing it to independently adjust each shading component according to the specific lighting scenario.

We evaluate \model{} on both synthetic and real-world datasets. It supports efficient point-light relighting under highly constrained training illuminations and significantly outperforms existing OLAT methods in OOD relighting tasks. We further demonstrate that both meta-learning and the differentiable Phong model individually contribute to improved generalization. Notably, \model{} generalizes well to unseen environment maps, despite being trained exclusively in OLAT scenarios.

\section{Preliminaries}
% \label{sec:preliminary}

\paragraph{Definition and challenges of OLAT.}
% \label{sec:problem_setup}

OLAT refers to a specialized setting for 3D reconstruction and relighting.
This involves illuminating the subject with a single point light in sequential exposures, capturing one image at each light position.
%
% In OLAT setting, the scene is lit by a moving point light and the training data is captured by a monocular camera, as shown in \figref{sections/fig:olat}.
%
Each capture is represented as a tuple $\{O_t, V_t, P_t\}$, where $O_t$ denotes the observed image, $V_t$ represents the camera configuration, and $P_t$ contains lighting information. In our context, $P_t$ specifies the position of the point light.
%
% In our work, \model{} is trained under a set of randomly captured images $\{I_t\}_{1:T}$. 
% Each image is grouped with its camera parameters and point-light position. 
%
Unlike the ``\textit{multiple lights, multiple cameras}'' setup, which requires multi-view images captured by multiple time-synchronized cameras for each lighting condition, OLAT relies on a single camera. This significantly simplifies data collection, particularly in scenarios with dynamic lighting.
%
% Despite the efficiency of data acquisition, 
The rapidly changing light positions in OLAT introduce a new challenge for 3D reconstruction. The color of a 3D point can vary significantly, increasing ambiguity for estimating true object geometries and illumination properties.

\vspace{-8pt}
\paragraph{Preliminary findings.}
\label{sec:preliminary}
We summarize existing OLAT relighting methods in \tabref{tab:olat_methods}.
% This paper focuses on relighting methods within the scope of 3DGS, due to its advantages in explicitly representing the geometric nature of the 3D scenes. \yb{not related}
%
% We have two empirical findings in the preliminary experiments.
%
% First, we witness significant degeneration in the performance of existing 3DGS-based relighting techniques~\cite{jiang2023gaussianshader} under the challenging OLAT training setup (please refer to Sec. \ref{sec:exp_nvs} for details).\hym{BiGS, RNG and GS-IR can handle olat setting}
%
% Second, we also evaluate NRHints~\cite{zeng2023relighting}, 
%
In the preliminary experiments, we evaluate NRHints~\cite{zeng2023relighting}, the prior art in NeRF-based relighting models. We observe that it underperforms in testing scenarios with out-of-distribution (OOD) lighting positions, a feature that we believe should be crucial for relighting methods.
Specifically, in \figref{sections/fig:nrhints}, the lighting in the training set is arranged on one side of the hemisphere, while the lighting in the test set is arranged on the opposite side (cameras are located on both sides). 
In such cases, NRHints fails to generate reasonable rendering results, likely due to its implicit modeling of shadows and specular reflections.
% In other words, inputting the point light source position into the network design limits its ability to generalize to more complex, OOD lighting conditions. 
%
We additionally evaluate a concurrent 3DGS-based approach~\cite{bi2024gs} and observe similar degraded results under the OOD relighting setup. 
% These findings highlight that OLAT remains a challenging task, as the simultaneous changes in lighting and viewing directions complicate the accurate learning of object intrinsic illumination properties.
These findings highlight that understanding intrinsic illumination properties under arbitrary lighting variations remains a challenging task, as the model tends to overfit to perspective-constrained observations and fail to leverage enough physical principles when dealing with unseen lighting distribution.

\begin{table}[t]
\centering
\caption{
Comparison of the OLAT relighting methods. 
%
% Bi-GS~\cite{liu2024bigs} and PRTGaussian~\cite{zhang2024prtgaussian} actually utilize intensive multi-light-multi-camera for training, and we include them as exceptions for they also use varying illumination for training.
}
\setlength\tabcolsep{8pt}
% \small
% \resizebox{0.7\linewidth}{!}{
\vspace{-5pt}
\begin{tabular}{lccccccc}
\toprule
                    &\cite{zhang2022iron}      % IRON  
                    & \cite{cheng2023wildlight} % WildLight
                    & \cite{fan2024rng}         % RNG
                    & \cite{zeng2023relighting} % NRHints
                    & \cite{bi2024gs}           % GS^3
                    & \cite{kuang2024olat}      % OLAT Gaussians
                    & Ours \\ \midrule
                    
                    %WildLight       RNG             NRHints         GS^3            OLAT Gaussians  IRON            Ours
    Point light     & \checkmark    & \checkmark    & \checkmark    & \checkmark    & \checkmark    & \checkmark    & \checkmark \\
    % Generalization to environment maps 
    %                 & $\times$      & $\times$      & $\times$      & $\times$      & $\times$      & $\times$      & \checkmark \\
    Shadow computation
                    & $\times$      & $\times$      & \checkmark    & \checkmark    & \checkmark    & \checkmark    & \checkmark \\
    OOD light       & $\times$      & $\times$      & $\times$      & $\times$      & $\times$      & $\times$      & \checkmark \\
    % Data reqt.  & \makecell{Single-point-light images \\ + all-light-on images}
    %             & \makecell{Single-point-light \\ images}
    %             & \multicolumn{7}{c}{\makecell{Single-point-light images}} \\
\bottomrule
\label{tab:olat_methods}
\end{tabular}
\vspace{-15pt}
\end{table}

\section{Method}
\label{sec:method}

In this section, we present the details of \model{} for addressing the OOD OLAT learning challenge:
\vspace{-5pt}
\begin{itemize}[leftmargin=*]
\vspace{-1pt}\item Sec. \ref{sec:phong}: We incorporate physical shading priors within Gaussian splatting via a differentiable Phong reflection model to decouple mixed illumination components.
\vspace{-1pt}\item Sec. \ref{sec:meta}: We introduce a bilevel optimization scheme to estimate light-independent scene geometries and intrinsic illumination properties, marking an early effort in volume rendering.
\vspace{-1pt}\item Sec. \ref{sec:optim}: We discuss the entire training pipeline and the implementation details.
\vspace{-3pt}
\end{itemize}

% \subsection{Gaussian-Phong Model}
\subsection{Differentiable Phong Model}
\label{sec:phong}

% We select the Blinn-Phong model to provide the illumination decomposition priors to 3DGS due to its simplicity and computational efficiency.

Our \textit{\model{}} method leverages simple yet generalizable physical priors from the Blinn-Phong model~\cite{phong1998illumination}, which captures three fundamental components of light transport: ambient, diffuse, and specular reflections.
The core idea is to disentangle these illumination components by learning the interactions between (i) the normal vectors of the Gaussian points, (ii) the viewing directions, and (iii) the ray direction from the point light. 

% Specifically, the ambient shading component represents the constant illumination presented in the environment, simulating how light scatters and reflects off other surfaces to establish a baseline brightness level.
Specifically, the ambient component represents constant environmental illumination, simulating indirect scattering from surrounding surfaces to establish a baseline brightness level.
The diffuse component, based on Lambertian law, describes light scattering in multiple directions from rough surfaces. %This component is calculated based on the angle between the light direction and the surface normal, ensuring that surfaces facing the light source appear brighter.
The specular reflection is computed based on the angle between the light direction and the bisector ($\mathbf{h}$) between the viewing direction ($\mathbf{v}$) and the light direction ($\mathbf{l}$), with a shininess exponent that represents different degrees of glossiness. 
%
% In all, the model can be described as a function $F: \mathbb{R}^3 \rightarrow \mathbb R$ that takes a 3D point $\mathbf x: (x,y,z)$ as input and returns a real number which is the associated light intensity. 
% In all, the model can be described as a function $F: \mathbb{R}^3 \rightarrow \mathbb {R}^3$ that takes a 3D point $\mathbf x: (x,y,z)$ as input and returns the associated illumination color. 

\begin{figure*}[t]
    \centering
    \includegraphics[width=\textwidth]{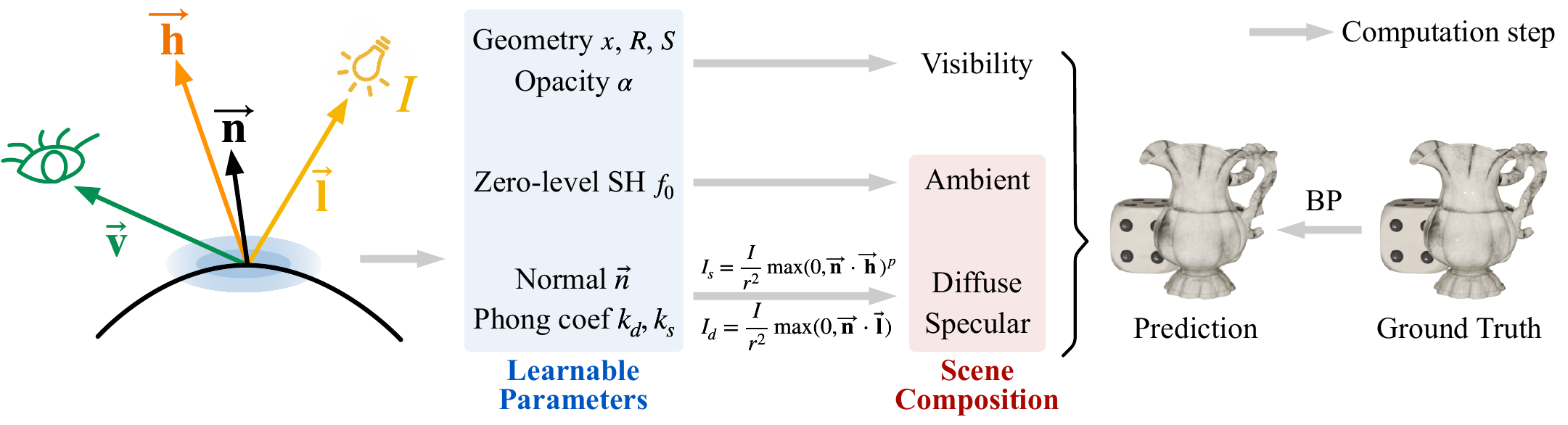}
    \vspace{-15pt}
    \caption{\textit{The model design of \model{}:} Our model decomposes the illumination effects by interacting the learned Gaussian points with rays originating from both the viewer and the light source. 
   }
    \label{sections/fig:model}
    \vspace{-5pt}
\end{figure*}

In \model{}, we extend the original 3DGS by introducing the decoupled computation for different shading components. 
As shown in \figref{sections/fig:model}, in addition to the basic Gaussian attributes (\textit{e.g.}, position $\mathbf x$, rotation $R$, scale $S$, opacity $\alpha$, and spherical harmonic coefficients $f$), each Gaussian point is further associated with a normal vector $\mathbf{n}$, a $3$-channel diffuse color ${k}_d$, and a $1$-channel specular coefficient $k_s$.  
The newly added attributes enhance the model's understanding of lighting effects, facilitating the learning process of light-independent object geometries.
The overall color of a Gaussian point is determined by:
\begin{equation}
\label{eq:phong}
    L_p = L_a + L_d + L_s=L_a + \displaystyle \sum \limits_{\lvert \text{lights}\lvert } (k_dI_d + k_sI_s),
\end{equation}
where $k_{\{d,s\}}$ and $I_{\{d,s\}}$ are the colors and intensities of the diffuse and specular components. $\lvert \text{lights}\lvert=1$ in OLAT.
% For each training view, we successively model the ambient, diffuse, and specular colors.
%
% The ambient color adds constant color to account for disregarded illumination and fills in black shadows.
%
In graphics, zero-order spherical harmonics (SH) coefficients typically represent the basic, uniform component of a function defined over the sphere, essentially the average or constant part of a lighting environment.
Therefore, we restrict Gaussians' SH coefficients to zero order, denoted by $f_0$, corresponding to the ambient color.
We compute the diffuse and specular light transports by multiplying the color and intensity of each component, where the intensities $I_d$ and $I_s$ are defined as:
\begin{equation}
I_d=\frac{I}{r^2}\max(0,\mathbf{n} \cdot \mathbf{l}), \quad I_s=\frac{I}{r^2}\max(0,\mathbf{n} \cdot \mathbf{h})^p,
\end{equation}
where $I$ denotes the light emitted intensity, which is a global learnable parameter, $\mathbf l$ denotes the point-to-light normalized vector, and $\mathbf{h}=\frac{\mathbf{v}+\mathbf{l}}{||\mathbf{v}+\mathbf{l}||}$ denotes the bisector of the point-to-camera normalized vector $\mathbf{v}$ and $\mathbf{l}$. 
We model the coefficients $k_d$ and $k_s$ implicitly, where $k_s$ is multiplied by the RGB color of the point light.
%Through this simple yet effective modeling, \model{} can decompose the scene lighting correctly.

\vspace{-8pt}
\paragraph{Shadow computation.}
We calculate shadow effects using a BVH-based ray tracing method~\cite{gao2023relightable}. 
%
% Ray-tracing calculates the received light intensity $T_i^\text{light}$ received by each point. 
%
Similar to the camera-to-point accumulated transmittance, the received light intensity $T_i^\text{light}$ represents the total transmittance of light along the light-to-point ray. It is therefore affected by the opacity of the surfaces encountered along this path.
For each point, we determine light visibility by tracing a ray from the Gaussian's center to the light source.
%
% The cumulative transmittance along this ray is given by $T_i^\text{light} = \prod_{j=1}^{i-1}(1-\alpha_j)$.
%
% To facilitate model convergence, we introduce a shadow coefficient $\phi$ shared by all Gaussian points, which acts as a scaling factor for this term.
%
Since only the diffuse and specular colors are influenced by incident light intensity, we update the color of each point by incorporating the light visibility factor into the diffuse and specular terms from Eq. \eqref{eq:phong}:
% \begin{equation}
% \label{eq:phong_shadow}
%     L_p = L_a + \phi \ T_i^\text{light} \sum\limits_{\lvert \text{lights}\lvert } (k_dI_d + k_sI_s).
% \end{equation}
\begin{equation}
\label{eq:phong_shadow}
    L_p = L_a + \ T_i^\text{light} \sum\limits_{\lvert \text{lights}\lvert } (k_dI_d + k_sI_s).
\end{equation}
This explicit formulation accounts for physics-based shadowing effects, provides strong interpretability, and improves generalizability compared to the implicit, high-dimensional shadow modeling methods.
However, integrating this module directly into \model{}'s pipeline presents challenges.
Under OLAT conditions, learning coherent geometry is challenging, leading to difficulties in accurately separating shadows from the scene. This, in turn, results in erroneous color predictions.
The following meta-learning scheme promotes the mutual learning of scene geometry and appearance, alleviating the difficulties in learning coherent attributes under varying illuminations.

\subsection{Meta-Learned Gaussian Relighting}
% \subsection{Meta-Learning of Light-Independent Geometries}
% \subsection{Bilevel Meta-Learning Shadow Integration}
\label{sec:meta}

% The primary challenge in OLAT relighting is learning light-independent geometries, due to the significant variations in appearance at the same spatial locations across different frames, causing the model to overfit to the training corpus with .

Existing 3D relighting methods exhibit performance degradation when handling out-of-distribution relighting, primarily due to overfitting lighting patterns to perspective-constrained observations, resulting in producing unreasonable lighting components (such as wrong specular and shadows).

% The primary challenge in OLAT relighting is learning light-independent geometries, due to the significant variations in appearance at the same spatial locations across different frames.
% %
% In our model, shading components such as diffuse color, specular color, and shadows are heavily influenced by the predicted normals of the Gaussian points.

%
To address this, we introduce a meta-learning framework based on bilevel optimization, which has been shown to effectively bridge the distribution shift between the training and testing domains, facilitating the generalization of optimized variables to unseen scenarios~\cite{chen2020closer}. This training strategy can also promote coherent geometry learning in complex OLAT tasks, as shown in the ablation study.
In \model{}, the intuition of incorporating bilevel gradient update is to mitigate overfitting to specific light conditions by explicitly simulating test samples with OOD light sources during each gradient update, thereby improving the model’s ability to generalize to varied lighting scenarios.

% This ``egg and chicken'' problem of interdependent optimization of the illumination coefficients and object geometries prevents the model from learning the correct scene properties, making it prone to local optima. 

As illustrated in Alg. \ref{algo:overall}, we organize the training processes with different $\{O_t, V_t, P_t\}$ as multiple learning tasks.
We alternate model training between these tasks, validating the optimized variables under one lighting condition using data sampled from other conditions. This learning procedure encourages both the lighting attributes ($f,k_a,k_s$) and the geometric attributes ($\mathbf{x},\mathbf{n},R,S,\alpha$) of the Gaussian points to converge cohesively.

To simulate test-time conditions, the training data is divided into support (\textit{training}) and query (\textit{validation}) sets. 
At each iteration, $2m$ samples are drawn from the training data to form the support set $\mathcal{D}_{1:m}^\text{sup}$ and query set $\mathcal{D}_{1:m}^\text{query}$ (Line 5).
Each support sample pairs with a query sample, denoted by subscript $i$.
During each training step, the model alternates between these tasks in the inner optimization loop (Lines 6-9) and the outer loop (Line 10).
% In the inner loop, the model is trained on each support set to generate hypothetical estimates $\theta_i^\prime$ and $\phi_i^\prime$, which are then evaluated on the query set in the outer loop to compute the final gradients and update the Gaussian variables.

\vspace{-8pt}
\paragraph{Inner optimization loop.} 
In the inner loop, the model is trained on each support set to independently generate $m$ sub-models, representing the hypothetical estimates $\theta_i^\prime$ under each lighting condition.
All sub-models start with the same parameters of $\theta$. % and $\phi$. 
%To simplify the learning process, we separate the training of Phong-related Gaussian parameters from the shadow coefficient.
%Specifically, in Line 7, we first compute $\theta^\prime$ based on the shadow-free Phong model in Eq. \eqref{eq:phong}. 
%
The light intensity $I$ is a global parameter that is also learnable and optimized jointly with the Gaussian attributes. It is omitted here for clarity.
%In Line 8, the hypothetical Gaussian attributes $\theta^\prime$ are then used in the shadow-on objective function in Eq. \eqref{eq:phong_shadow} to optimize the shadow coefficient and obtain $\phi_i^\prime$.
%
We specify the loss function $\mathcal{L}_\text{final}$ in subsequent Sec. \ref{sec:optim}.

\vspace{-8pt}
\paragraph{Outer optimization loop.}
In the outer loop, a global gradient update is performed across all sampled tasks, updating the Gaussian attributes.
% where both the Gaussian attributes and shadow coefficients are collectively updated.
% 
We first compute the loss function for each query sample, $\mathcal{L}_\text{final}(\cdot, \cdot; \mathcal D_i^\text{sup})$, using the corresponding inner-loop model hypotheses $\theta_i^\prime$.% and $\phi_i^\prime$.
We then aggregate the $m$ losses to update the initial $\theta$.% and $\phi$.
This approach, involving task-specific adaptation followed by a global update, enables the model to learn generalizable representations across varying lighting conditions.

\begin{algorithm}[t]
    \caption{Meta-Training for OOD Relighting}
    \label{algo:overall}
    \begin{algorithmic}[1]
    \small
    \State \textbf{Input: } Training set $ \{\mathcal D_t\}_{1:T}$, where $ D_t= \{$light position, camera parameters, RGB image$\}$
    \State \textbf{Hyperparameters:} Learning rates $\alpha, \beta$
    \State \textbf{Initialized parameters:} Pretrained Gaussian attributes $\{\theta_k\}$, where $\theta_k=(\mathbf{x},\mathbf{n},R,S,\alpha,f_0,k_d,k_s)$%, shadow coefficient $\phi$
    % light intensity $I$,
    \While{not converge} 
        % \State \textit{\# Inner loop:}
        \State Sample disjoint data $\mathcal D_{1:m}^\text{sup}$ and $\mathcal D_{1:m}^\text{query}$ from $\{\mathcal D_t\}_{1:T}$ % \Comment{Using $m$ tasks for one iteration}
        \For{task $i$ in $\{1:m\}$}  \Comment{\texttt{Inner optimization loop}}
            % \State $\theta_i^\prime \leftarrow \theta - \alpha_1 \nabla_\theta\mathcal{L}_\text{final}(\theta; \mathcal D_i^\text{sup})  $ \Comment{\texttt{Shadow off}}
            % \State $\phi_i^\prime \leftarrow \phi - \alpha_2 \nabla_\phi\mathcal{L}_\text{final}(\theta_i^\prime, \phi; \mathcal D_i^\text{sup})  $ \Comment{\texttt{Shadow on}}
            \State $\theta_i^\prime \leftarrow \theta - \alpha \nabla_\theta\mathcal{L}_\text{final}(\theta; \mathcal D_i^\text{sup})  $
            % \State $\phi_i^\prime \leftarrow \phi - \alpha \nabla_\phi\mathcal{L}_\text{final}(\theta, \phi; \mathcal D_i^\text{sup})  $
        \EndFor
        % \State \textit{\# Outer loop:}
        % \State $\{\theta,\phi\} \leftarrow \{\theta,\phi\} - \beta \sum_{i=1}^{m}\nabla_{\theta,\phi} \mathcal{L}_\text{final}(\theta_i^\prime, \phi_i^\prime; \mathcal D_i^\text{query})  $\ \Comment{\texttt{Outer optimization step}}
        \State $\{\theta\} \leftarrow \{\theta\} - \beta \sum_{i=1}^{m}\nabla_{\theta} \mathcal{L}_\text{final}(\theta_i^\prime; \mathcal D_i^\text{query})  $\ \Comment{\texttt{Outer optimization step}}

    \EndWhile
\end{algorithmic}
\end{algorithm}

% \subsection{Optimization}
% \subsubsection{Training Strategy}
\subsection{Entire Training Pipeline}
\label{sec:optim}

We employ a three-stage training scheme.
The first stage, similar to 3DGS, focuses on training the core Gaussian attributes, including position $\mathbf x$, rotation $R$, scale $S$, opacity $\alpha$, and spherical harmonic coefficients $f$.
%
% Experimental results indicate that, a
After this stage, the model tends to capture a rough geometry and an average color across different illuminations, serving as an effective initialization for further refinement.
In the second stage, we incorporate the normal attributes into the optimization process, acknowledging that learning a Phong model heavily relies on accurate normal estimations.
In the final stage, we integrate the diffuse and specular components, training all parameters concurrently through meta-learning.

\vspace{-8pt}
\paragraph{Objective functions.}
We follow 3DGS~\cite{kerbl20233d} to compute the deviation of the predicted image from the ground truth, using an L1 regularization and a D-SSIM term as the RGB loss.
We also employ a sparse loss~\cite{jiang2023gaussianshader} to encourage the opacity values $\alpha$  of the Gaussian spheres to approach either $0$ or $1$, thereby facilitating learning with opaque objects.
For normal estimation, we follow GaussianShader~\cite{jiang2023gaussianshader} to compute the normal vectors using Gaussian's shortest axis direction, and progressively align them with the depth-inferred pseudo-normals.

\vspace{-8pt}
\paragraph{Implementation details.}
We implement \model{} using PyTorch~\cite{paszke2019pytorch}. For optimization, we use the Adam optimizer~\cite{kingma2014adam} with the same parameters as those specified by \cite{kerbl20233d}. 
%kerbl20233d
A consistent set of hyperparameters is applied across all scenes, with details provided in the Appendix.
%
% Specifically, we use a standard exponential decay scheduling technique for the diffuse color prior to facilitate smooth training. We exponentially decay the diffuse color loss from $0.02$ to $0.002$ during the first $1k$ epochs in meta-learning stage, and fix it in the rest of the training.
%
Our training procedure contains three stages: the first two stages each last for $4k$ iterations, followed by the final meta-learning stage of $5k$ iterations.
The entire training process requires roughly one hour on a single NVIDIA RTX 3090 GPU.
%
%During inference, it achieves $20$ FPS for image synthesis.
%
% Compared to existing OLAT-based methods like NRHints, which require more than a day for training on the same hardware, our method has a significant time advantage, thanks to the efficiency of Gaussian splatting and effective meta-learning.
\section{Experiments}
\label{sec:exp}

% In this section, we present 
% (i) qualitative and quantitative comparisons on both synthetic and real-world OLAT datasets with existing algorithms; 
% (ii) \model{}’s capability of jointly estimating the scene's geometry and inherent color by predicting lighting effects and removing shadows; 
% (iii) out-of-distribution relighting comparisons with baselines which showcases our model's generalization to unseen illumination distributions.
% (iiii) Ablation studies of \model{} to investigate the effectiveness of each part of our methods. 

% syn - NRHints, IRON and TensoIR
\begin{figure}[t]
    \centering
    \includegraphics[width=\columnwidth]{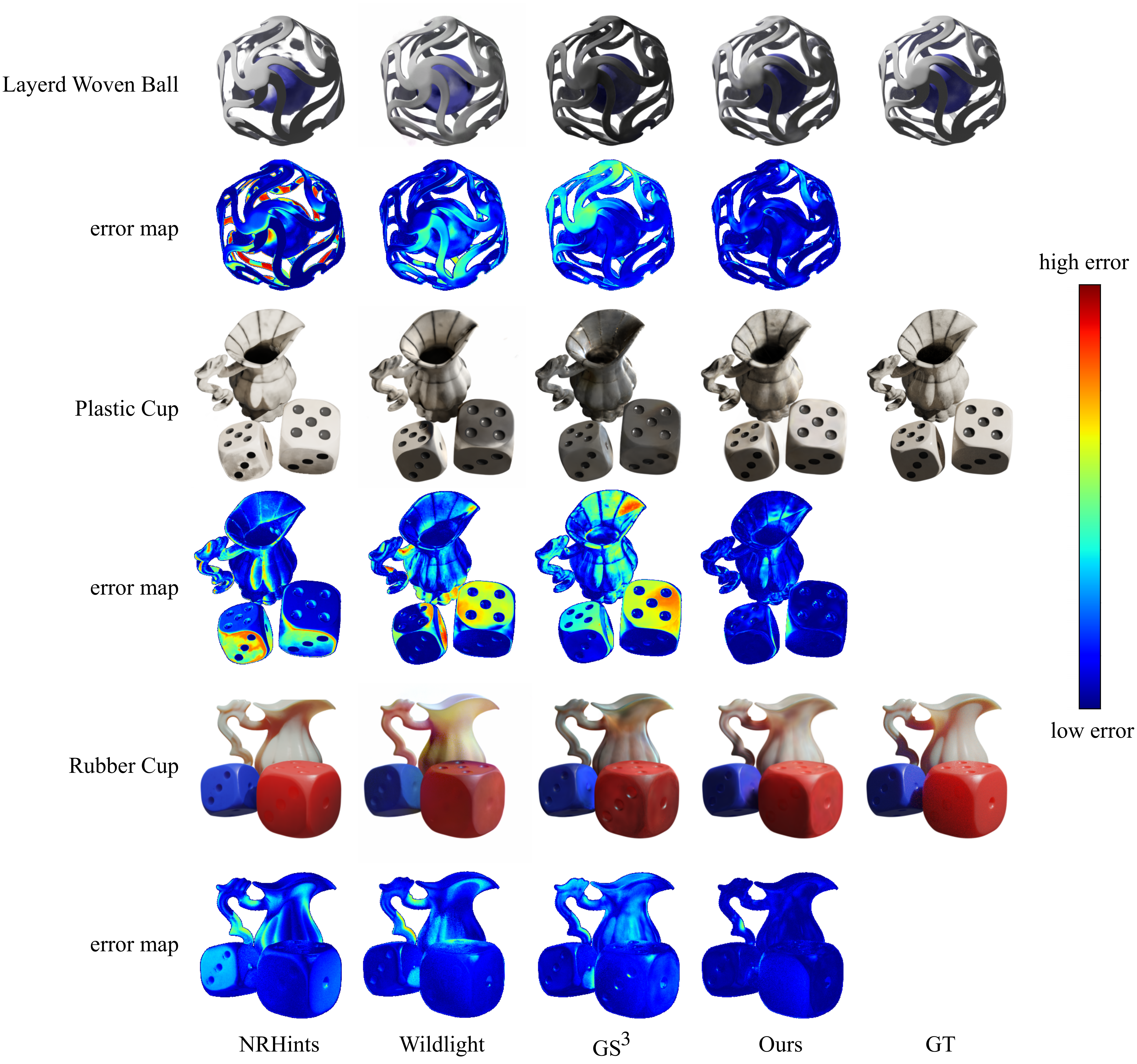}
    \vspace{-20pt}
    \caption{
    \textit{OOD relighting results on synthetic data:} 
    We present rendered novel views and error maps. While baselines often misrepresent shadows or light-dependent effects (\textit{e.g.}, incorrect shadows in \textit{Plastic Cup}), our model better infers surface appearance.
    }
    \label{sections/fig:ood_syn}
    \vspace{-10pt}
\end{figure}

\subsection{Experimental Settings} 
\label{exp_setting}

We perform three types of OLAT relighting experiments: %, enabling a comprehensive evaluation of the models beyond previous in-distribution novel view relighting:
\vspace{-5pt}
\begin{itemize}[leftmargin=*]
    
    \item\textit{OOD relighting:}
    We introduce a novel view synthesis setup to assess the generalizability of 3D relighting models under OOD lighting conditions. The point lights in the training set are positioned on one side of the upper hemisphere, while the lighting in the test set is placed on the opposite side. 
    % We also show the multi-light relighting results of the trained model.
    %
    % This configuration challenges the model to generalize beyond the lighting conditions it encountered during training.
    % To better validate the meta-learning generalization ability in OOD scenarios, we generate OOD versions of all scenes. Specifically, in this setup, the lighting in the training set is arranged on one side of the hemisphere, while the lighting in the test set is on the opposite side, with cameras positioned on both sides. 
    
    \vspace{-2pt}\item \textit{Camera-light-colocated relighting:}
    This setup mirrors the one used in IRON~\cite{zhang2022iron}, where the camera and point light are co-located for each image. 
    % Both the training and testing light positions are sampled from the same region, reflecting an in-distribution relighting scenario.
    %
    Similar to the OOD setting, the colocated configuration also represents a constrained lighting regime, as the light source is fixed to the camera pose. 
    For fair comparison, we generate colocated versions of the synthetic scenes accordingly.
    %
    % This experiment follows the same OLAT setup as in Novel View Relighting, except that in each image sample, the camera position is identical to the point light position. Given that IRON~\cite{zhang2022iron} is specifically designed for camera-light-colocated scenes, we also generate the colocated version of synthetic scenes for a fair comparison. 
    % Both the training and testing cameras are randomly sampled from the same region, with each image being lit by a light source positioned at the same position as the camera.
    %
    % Cameras are located on both sides of the hemisphere.
    \vspace{-2pt}\item \textit{Environment map relighting:} We evaluate the generalization of the models exclusively trained in the OLAT setup to novel view synthesis with unseen environment lighting maps.
    \vspace{-3pt}
\end{itemize}

% realworld - NRHints, IRON
\begin{figure*}[t]
    \centering
    \includegraphics[width=\linewidth]{sections/fig/ood_realworld.pdf}
    \vspace{-20pt}
    \caption{
    \textit{OOD relighting results on real-world data:} 
    As highlighted with the red boxes, baseline models struggle with some level of global shading consistency, such as color shifts, wrong shadows, and floating artifacts.
    Our approach presents physically plausible specular highlights and geometrically consistent shadows that closely match ground truths.
    }
    \label{sections/fig:ood_realworld}
    \vspace{-15pt}
\end{figure*}

\begin{table*}[t]
\centering
\vspace{-3pt}
\caption{\textit{PSNR results for OOD relighting.} See our Appendix for full comparisons.
% Full comparisons in all evaluation metrics are included in the \textit{supplementary materials}. 
% $^\dagger$We evaluate the time cost on four RTX 4090 GPUs for NRHints and one 3090 GPU for the others.
}
\vspace{3pt}
\setlength\tabcolsep{3pt}
\small
% \resizebox{1.0\linewidth}{!}{
\begin{tabular}{l ccc ccccccc}%|cc}
\toprule
\multirow{2.5}{*}{Method} & \multicolumn{3}{c}{Synthetic} & \multicolumn{7}{c}{Real-world} \\%& \multicolumn{2}{|c}{Efficiency$^\dagger$}
\cmidrule(r){2-4}\cmidrule(r){5-11}
                            & Ball & PlasCup & RubCup
                            & Cat & Catsmall & CupFabric & Fish & FurScene & Pikachu & Pixiu \\% & Train & Test-FPS \\
\midrule
NRHints~\cite{zeng2023relighting}       &  17.25&23.92&27.44    &  18.04&24.63&24.65&22.57&21.55&24.00&23.03 \\%&   16 hours&1/150    \\ 
WildLight~\cite{cheng2023wildlight}     &  21.73&20.95&24.02    &  18.65&22.53&24.03&21.47&20.33&19.09&20.22 \\%&   16 hours&1/120    \\
GS$^3$~\cite{bi2024gs}                  &  18.84&20.30&24.37    &  17.66&23.34&25.04&21.12&17.34&24.11&19.63 \\%&   1.5 hour&25       \\
% PRTGaussian~\cite{zhang2024prtgaussian} &  14.56&14.95&23.98    &  12.67&20.37&18.41&19.72&17.42&16.21&15.42\\
Ours &  \textbf{26.76 }&\textbf{27.54}&\textbf{27.95}  &  \textbf{26.45}&\textbf{26.44}&\textbf{27.29}&\textbf{24.68}&\textbf{24.82}&\textbf{25.54}&\textbf{25.65} \\%&1 hour&20 \\ 
\bottomrule
\end{tabular}
% }
\label{tab:ood_syn}
\vspace{-10pt}
\end{table*}

We evaluate \model{} on $3$ synthetic scenes and $7$ real captured scenes from NRHints~\cite{zeng2023relighting}, featuring a diverse range of materials, complex object shapes, significant self-occlusion, and intricate shadow effects. Each image is rendered (or captured) with a unique camera pose and point-light position. Full details of the datasets are provided in the Appendix.

For each synthetic scene, we generate a total of $600$ OLAT images using Blender, with $500$ for training and $100$ for testing, following out-of-distribution or colocated data patterns. 
% For the camera-light-colocated experiment, we recreate these scenes to align with the colocated setup.
%
For the real-world scenes, we split the data according to their point light positions to construct out-of-distribution datasets, and use 600 training images per scene---a significantly smaller subset than the full NRHints dataset.

% \paragraph{Baselines and metrics.}
We compare \model{} against both NeRF-based and Gaussian-based models, with a particular focus on the OOD relighting setup. The baseline models include state-of-the-art approaches from the past two years: GS$^3$~\cite{bi2024gs}, NRHints~\cite{zeng2023relighting}, WildLight~\cite{cheng2023wildlight}, Relightable 3DGS ~\cite{gao2023relightable}, GaussianShader ~\cite{jiang2023gaussianshader}, and IRON~\cite{zhang2022iron}.
The evaluation metrics include PSNR, SSIM~\cite{wang2004image}, and LPIPS~\cite{zhang2018unreasonable}.

% For OOD relighting, we compare with three OLAT baselines: NRHints~\cite{zeng2023relighting}, WildLight~\cite{cheng2023wildlight}, and GS$^3$~\cite{bi2024gs}. 
% %
% For in-distribution relighting, we mainly with 3DGS ~\cite{kerbl20233d}, Relightable 3DGS ~\cite{gao2023relightable}, and GaussianShader ~\cite{jiang2023gaussianshader}.
% %
% Specifically, we compare with IRON~\cite{zhang2022iron} under the camera-light-colocated setup.
%
% We train 3D Gaussian Splatting, GaussianShader and Relightable 3DGS for 30k epochs, 30k epochs and 70k epochs respectively according to their original settings. 

\begin{figure*}[t]
\vspace{10pt}
    \centering
    \includegraphics[width=0.99\linewidth]{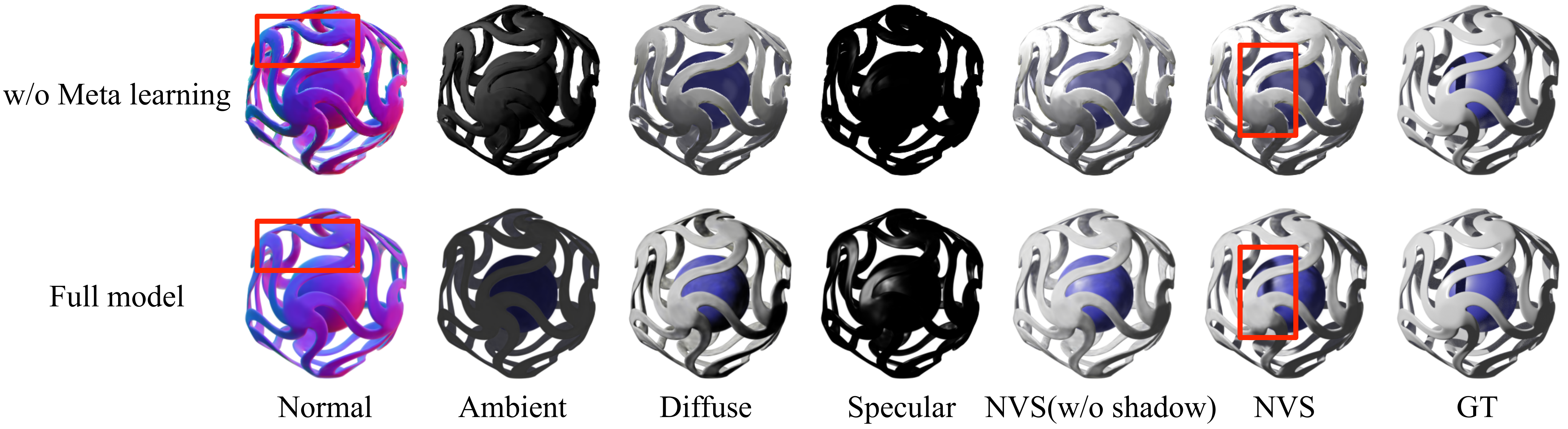}
    \vspace{-5pt}
    \caption{\textit{Ablation studies with qualitative results:} As shown in the decoupling components, the meta-learning scheme is essential for learning object illumination properties and geometries. Without it, the model generates plain lighting effects, particularly for the diffuse and specular components.
    % We omit the visualization of ``w/o scene component smooth loss'' and ``w/o visibility sparse loss'' as their absence leads to similar artifacts to those seen in ``w/o diffuse prior loss''.
    }
    \vspace{-5pt}
    \label{sections/fig:ablation}
\end{figure*}

\subsection{Results for Out-of-distribution OLAT Relighting}

We evaluate \model{} in scenarios with OOD point light positions to explore its generalizability to unseen illuminations. 
The qualitative and quantitative results are respectively presented in \figref{sections/fig:ood_syn}, \figref{sections/fig:ood_realworld}, and \tabref{tab:ood_syn}.
Our model remarkably outperforms the baselines in handling complex light interactions.
For example, in \figref{sections/fig:ood_realworld}, the lighting and shading effects on the cat’s fur are rendered with high fidelity, while baseline models tend to produce incorrect color tones.
% , indicating that they fail to model the illumination distribution correctly.
%
Furthermore, as indicated by the error maps in \figref{sections/fig:ood_syn}, all of three OLAT baselines tend to provide unrealistic results, especially for regions with significant specular reflections or shadows.
Please see our Appendix for video visualizations on OOD relighting, where baselines' performance drops as the light enters the OOD region, while our model provides better relighting results.

The generalizability of \model{} to OOD illumination stems from two key factors.
First, it is due to the Phong model’s ability to capture the underlying principles of light interaction.
For example, instead of simply viewing specular highlights as the view-dependent color of a point (which is not generalizable), our method explicitly calculates the interactions of rays and surface normals, enabling the model to generate accurate highlights even in areas not directly illuminated during training.
Second, meta-learning further enhances generalization by simulating test-time conditions during training, effectively reducing overfitting to specific illuminations.
In the Appendix, we empirically show that the benefit of meta-learning does not come from a larger batch size.

% During inference, in occluded regions, our model displays only the ambient color, avoiding the unrealistic highlights that NRHints tends to produce.
%
% Additionally, it calculates specular highlights based on surface normals, enabling it to generate accurate highlights even in areas not directly illuminated during training.
%
% However, for NeRF-based methods, treating the light position as an input to the implicit network limits their ability to generalize to more complex, OOD lighting conditions. 
%
% PRTGaussian: our method does not require extensive multi-light, multi-camera data collection~\cite{bi2024gs,zhang2024prtgaussian}, making it more practical for real-world relighting scenarios. % 而且超级过拟合 所有baseline都过拟合很严重

\vspace{-8pt}
\paragraph{Ablation studies.}

\begin{wraptable}{r}{0.5\linewidth}
\small
% \begin{table}[t]
\vspace{-18pt}
\caption{\textit{Ablation studies of each model component for OOD relighting.} We show the average results of all three synthetic scenes.}
\vspace{-10pt}
% \small
\begin{center}
\resizebox{1.0\linewidth}{!}{
    \begin{tabular}{lccc}
        \toprule
        Method & PSNR$^\uparrow$ & SSIM$^\uparrow$ & LPIPS$^\downarrow$ \\
        \midrule
        Full model &  \textbf{27.42}&\textbf{0.9546}&\textbf{0.0505}\\
        w/o Meta-learning& 19.14&0.8781&0.0892 \\
        w/o Shadow&  21.53&0.9105&0.0735 \\
        % w/o Smooth loss, Eq. \eqref{eq:smooth}&    25.39&0.9335&0.0628 \\
        % w/o Diffuse prior loss, Eq. \eqref{eq:diffuse}&     24.98&0.9242&0.0660 \\
        % w/o visibility sparse loss, Eq. \eqref{eq:sparse}& &&
        \bottomrule
    \end{tabular}
}
\label{tab:ablation_study}
\end{center}
\vspace{-10pt}
% \end{table}
\end{wraptable}
We conduct a series of ablation studies to validate the effectiveness of the meta-learning scheme as well as the shadow computing. From \tabref{tab:ablation_study} and \figref{sections/fig:ablation}, we observe that the proposed meta-learning training scheme strongly impacts performance. Without it, the model struggles to converge smoothly and make unrealistic component estimations, resulting in plain rendering quality.

% Qualitative and quantitative results are presented in \figref{sections/fig:ablation} and \tabref{tab:ablation_study}, respectively, showcasing the contribution of each model component.
%
% As shown in \tabref{tab:ablation_study} and \figref{sections/fig:ablation}, our regularization terms in the final objective function contribute to the final performance of the model.
%
% We omit the visualization of ``w/o scene component smooth loss'' and ``w/o visibility sparse loss'' as their absence leads to similar artifacts to those seen in ``w/o diffuse prior loss''.
% We omit the visualization of ``w/o Smooth loss'' as its absence leads to similar artifacts to those seen in ``w/o Diffuse prior loss''.
%
% We include further discussions on the ablation studies in the Appendix.

% \paragraph{Robustness on camera parameters.}
% NRHints network structure includes learnable camera parameter adjustments, while our model does not explicitly incorporate such adjustments. However, our method still outperformed NRHints in OOD relighting results, indicating that minor deviations in camera calibration do not hinder the correct convergence of our algorithm. 
% Overall, these results highlight the robustness and efficacy of \model{} in handling complex relighting tasks and achieving superior performance in scene decomposition.

\subsection{Generalization to Other Relighting Setups with Constrained OLAT Training}

\begin{figure}[t]
  \centering
  \includegraphics[width=\linewidth]{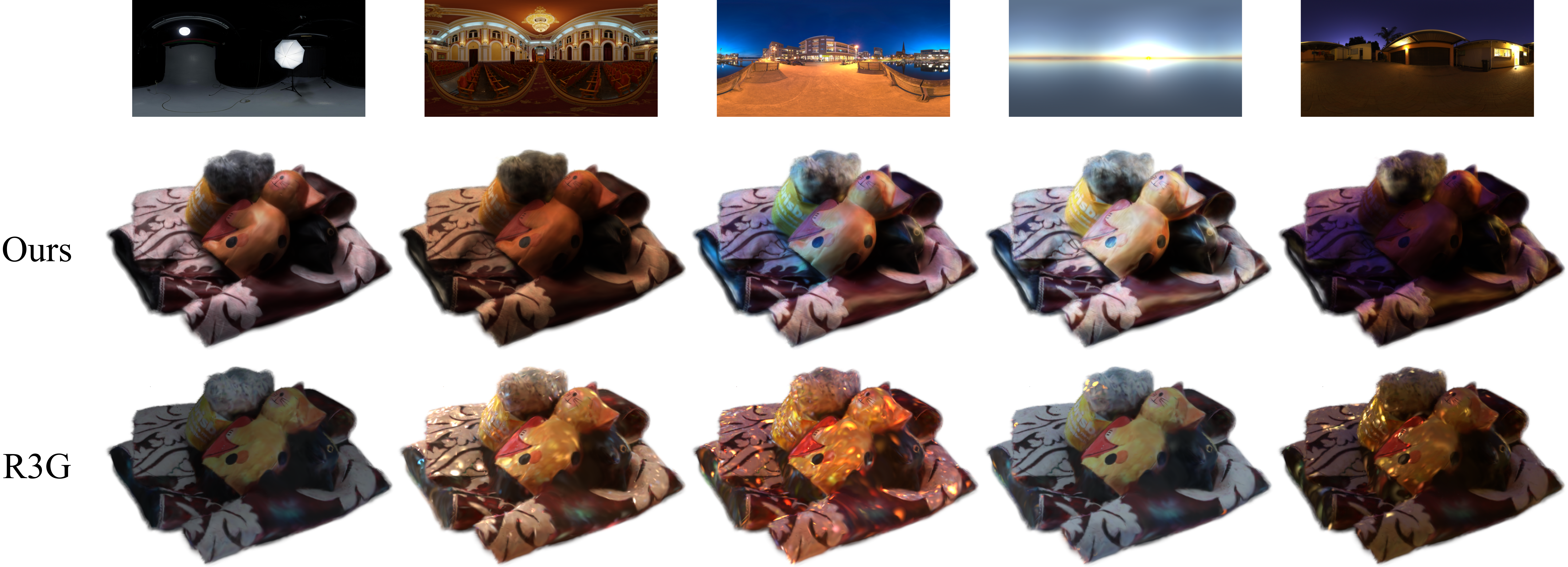}
  \vspace{-15pt}
   \caption{\textit{Relighting with environment maps:} Our method, trained under OLAT settings, successfully generalizes to unseen environmental lighting; while the compared method is trained in an all-light-on setup that provides multiple lighting conditions per training view, it still fails to generalize to perform relighting with novel environment maps and exhibits visible artifacts.}
   \vspace{-5pt}
   \label{fig:envmap}
\end{figure}

\paragraph{Camera-light-colocated relighting.}

\begin{wraptable}{r}{0.38\linewidth}
\small
\vspace{-18pt}
% \begin{table}[t]
\caption{\textit{Novel view synthesis results in PSNR under camera-light-colocated relighting setup.} Full results in all evaluation metrics are included in the Appendix.}
\vspace{-10pt}
\begin{center}
% \scalebox{0.9}{
% \small
% \setlength\tabcolsep{15pt}
\resizebox{1.0\linewidth}{!}{
    \begin{tabular}{lccccccccc}
        \toprule
        Method & Ball & PlaCup & RubCup\\
        \midrule
        IRON~\cite{zhang2022iron}  &  26.99&34.43&36.22\\
        Ours   &   \textbf{38.72}&\textbf{36.90}&\textbf{38.89} \\
        \bottomrule
    \end{tabular}
% }
}
\end{center}
\label{tab:syn_iron_psnr}
\vspace{-10pt}
% \end{table}
\end{wraptable}
We evaluate \model{} in the camera-light-colocated setup, as introduced by IRON~\cite{zhang2022iron}. 
%
% Because the illumination is tied to the camera’s position throughout training, this setting---similar to the OOD case---offers a narrow range of lighting conditions for supervision.
%
Like the OOD scenario, the colocated setup also imposes limited lighting diversity, as the point light is tied to the camera’s position throughout training.
Quantitative results in \tabref{tab:syn_iron_psnr} show that \model{} outperforms IRON, achieving more accurate illumination inference under new point positions, and generating fewer rendering artifacts. Visualizations corresponding to these results can be found in the Appendix.

\vspace{-8pt}
\paragraph{Environment map generalization.}
Our model generalizes effectively to environment map relighting.
We approximate global illumination by importance sampling point lights from the environment map based on their pixel intensities. 
Notably, for our model in particular (unlike the compared methods), the sampled light directions at test time may follow a distribution different from those encountered during training, presenting a significant challenge for OOD generalization.
Instead of assuming infinite light source distance, we simulate distant lighting by placing point lights along environment directions at a fixed distance (twice the scene radius).
We compare our method with R3G~\cite{gao2023relightable} and GaussianShader~\cite{jiang2023gaussianshader}. 
Both methods fail to converge when trained directly on out-of-distribution (OOD) data, so we construct training data that covers the full hemispherical illumination space.
Even with this setup, GaussianShader still fails to converge, while Relightable 3DGS struggles with the OLAT learning task and exhibits noticeable artifacts.
As shown in \figref{fig:envmap}, \model{} produces diverse and visually plausible relighting results under complex environment maps, demonstrating its robustness and generalization capability.

% As shown in \figref{fig:envmap}, despite being trained on limited out-of-distribution point-light data, our model produces visually plausible relighting results under complex environment maps.

\vspace{-8pt}
\paragraph{More results.}
To validate \model{}'s capability in handling complex lighting conditions, we conduct experiments with test sets containing $2$--$3$ light sources. We also present the results for free-viewpoint relighting with in-distribution point light positions. Please refer to the Appendix for details.

% \paragraph{Efficiency.} 
% As shown in Table \ref{tab:ood_syn}, \model{} significantly outperforms NRHints in efficiency. It renders an image at $20$ FPS on an RTX 3090 GPU, while NRHints takes $150$ seconds per image using four RTX 4090 GPUs. 

% \newpage
\section{Related Work}
% \vspace{-5pt}

% \paragraph{Learning-based rendering.}

% \subsection{Learning-Based Volume Rendering}
% Volume rendering is a visualization technique that converts 3D datasets into 2D images and is widely used in fields such as medical imaging, scientific visualization, and computational fluid dynamics. 
%
Recent differentiable volume rendering techniques, including NeRF-based~\cite{mildenhall2021nerf}, SDF-based~\cite{yariv2021volume,yu2022monosdf}, and 3DGS-based~\cite{kerbl20233d} methods, have significantly improved the quality and efficiency of novel view synthesis for 3D scenes.
NeRF-based methods utilize deep neural networks to model volumetric scene functions, encoding color and density to synthesize high-quality images from sparse viewpoints~\cite{mildenhall2021nerf,liu2020neural, zhang2020nerf++}.
However, these approaches typically require substantial training time. 3DGS significantly reduces both training and inference time. It converts point cloud data into a continuous volumetric representation by applying Gaussian kernels to point cloud data, facilitating rendering and further processing. 
This technique is now widely adopted for efficient 3D and 4D reconstructions across varied data types~\cite{wu2024deferredgs,jiang2023gaussianshader,luiten2023dynamic,wu20234d,yang2023deformable}.
%
% While not inherently a learning-based method, recent implementations~\cite{kerbl20233d} have incorporated machine learning to optimize kernel parameters dynamically, adapting to specific data characteristics or desired output quality. Such adaptive splatting techniques are increasingly used to enhance the efficiency of 3D and 4D reconstruction~\cite{wu2024deferredgs,jiang2023gaussianshader,luiten2023dynamic,wu20234d,yang2023deformable}, demonstrating significant potential to handle diverse data types and visualization requirements within short time.
%
%Among various learning-based volume rendering techniques, 3D Gaussian Splatting has gained particular attention for its practical efficiency and advanced reconstruction quality. 

% \vspace{-8pt}
% \paragraph{3D relighting.}

The task of 3D relighting involves altering the illumination in a 3D scene while maintaining its geometry. It requires decomposing materials and lighting of a scene from multiple images, which is challenging due to its high-dimensional nature.
Recent advances in volume rendering have introduced various solutions for this task, including those based on neural fields~\cite{zhang2021nerfactor,wu2023nerf,sun2023sol,sarkar2023litnerf,Srinivasan_2021_CVPR,liang2023envidr,Toschi_2023_CVPR,sun2023sol,Yang_2023_CVPR} and the methods based on Gaussian point clouds~\cite{wu2024deferredgs,jiang2023gaussianshader,gao2023relightable,Zhang_2021_CVPR}.
A limitation of these methods is that they require a substantial volume of multi-view images captured under individual lighting conditions, making them impractical in real-world scenarios.
%
% Relightable 3D Gaussian~\cite{gao2023relightable} introduces an innovative Gaussian-based visibility calculation algorithm, simulating ray-tracing effects and incorporating visibility data into the spherical harmonics parameters of each point, but struggles with OLAT settings. It integrate the shadow effect into the objects' color, therefore are effective only when the overall visibility (light occlusion correlations) of the scene remains constant. Also, its point-based representation poses challenges for effective shadow computation due to its discrete nature and lack of continuous surface information.
%
Several studies have proposed NeRF-based 3D relighting techniques within the OLAT framework~\cite{xu2023renerf,zhang2021nerfactor,zeng2023relighting}, which greatly reduce the demands on training data. 
Nonetheless, these techniques tend to be computationally intensive because of NeRF's inherent complexity in the volume rendering process.
In parallel, concurrent 3DGS-based approaches~\cite{fan2024rng,bi2024gs}  have also explored the OLAT relighting challenge. Despite their efforts, these approaches still face difficulties when dealing with lighting conditions not encountered during training, similar to the limitations observed in NeRF-based methods.

In contrast, our method addresses this issue by incorporating specific physical priors, represented by the Phong reflection model, into the Gaussian splatting framework. We have also designed a meta-learning approach to enhance the method's generalizability to OOD relighting scenarios.

% requires ground-truth environment maps for training, and shows less promising results on real-world scenes (with lower PSNR compared with 3DGS).
%
% NRHints~\cite{zeng2023relighting} and GS$^3$~\cite{bi2024gs} offer effective neural reflection models for OLAT, but struggle with unseen lighting conditions (as shown in \figref{sections/fig:nrhints}). 
%
% and multi-light scenarios due to its implicit modeling.
%
% NRHints~\cite{zeng2023relighting} offer effective neural reflection models for OLAT. Besides achieving accurate learning of scene geometry and material properties, it also provides more precise predictions of highlights and shadows compared to other methods, through the use of shadow hints and specular hints. However, it incorporates the point light position as a 3D input to the network, limiting the model to scenarios with only one point light source. These limitations are due to the implicit modeling nature that potentially limits the model's ability to generalize to more complex, out-of-distribution (OOD) lighting conditions.
%
% This explicit approach ensures better generalization and performance in diverse lighting conditions, addressing the limitations of previous methods.

\section{Conclusions and Limitations}
\label{sec:conclusion}

% To tackle the intertwined shadow learning problem inherent in OLAT, we integrate a meta-learning scheme into our training pipeline.
% %
% Our primary goal is to demonstrate that recent approaches, such as NRHints, lack the ability to generalize to new light positions. 
% %
% We addressed this limitation by using meta-learning in neural rendering, which enhances the model's decomposition learning ability. 
% %
% Our model may not perform as well as NRHints under non-OOD lighting positions, mainly in handling fine details, which is primarily due to its explicit representation. 
% %
% However, this representation contributes to our model's strong OOD generalization capability, enabling the model to effectively learn the logic of light decoupling. 
% %
% Additionally, our model is significantly faster than NRHints, thanks to the efficiency of Gaussian splatting and effective meta-learning.

% , a decoupled Gaussian Splatting method that decomposes the illumination of the scene while inferring its inherent geometry and color.
%
%Our model derives accurate geometry information and lighting variations from a set of captured images with rapidly changeable shadows.

In this paper, we explored a novel and challenging problem: out-of-distribution (OOD) 3D relighting.
To address this, we proposed \model{}, which builds on Gaussian splatting and presented a novel bilevel optimization-based meta-learning framework that explicitly promotes generalizable Gaussian geometry and appearance learning. This meta-learning formulation allows the model to adapt to varying light sources and viewpoints, even when the training data is biased or sparsely sampled in the lighting space.
Furthermore, \model{} incorporates a differentiable Blinn-Phong reflection model within Gaussian splatting, effectively disentangling lighting effects into ambient, diffuse, and specular shading components, thereby improving the physical realism and reconstruction fidelity under diverse lighting.
Extensive experiments on both synthetic and real-world datasets demonstrate that \model{} significantly outperforms existing OLAT relighting approaches under OOD relighting conditions. It also achieves strong generalization to complex environment maps.

Despite these advances, several limitations remain. An open challenge is ensuring the robustness of our approach under more complex lighting conditions. Moreover, our current framework only accounts for direct illumination; incorporating indirect lighting could further enhance the quality of relighting.
Additionally, the current model assumes simple Phong reflection, which may limit fidelity when modeling materials with strong subsurface scattering or anisotropic reflectance.

{
    \small
    \bibliographystyle{plain}
    \bibliography{main}

@String(CVPR= {IEEE Conf. Comput. Vis. Pattern Recog.})

@String(ICCV= {Int. Conf. Comput. Vis.})

@String(TOG= {ACM Trans. Graph.})

@String(CVPR  = {CVPR})

@String(ICCV  = {ICCV})

@String(TOG   = {ACM TOG})

@article{wu2024deferredgs,
  title={Deferred{GS}: Decoupled and Editable {G}aussian Splatting with Deferred Shading},
  author={Wu, Tong and Sun, Jia-Mu and Lai, Yu-Kun and Ma, Yuewen and Kobbelt, Leif and Gao, Lin},
  journal={arXiv preprint arXiv:2404.09412},
  year={2024}
}

@inproceedings{liang2023envidr,
  title={{ENVIDR}: Implicit differentiable renderer with neural environment lighting},
  author={Liang, Ruofan and Chen, Huiting and Li, Chunlin and Chen, Fan and Panneer, Selvakumar and Vijaykumar, Nandita},
  booktitle={ICCV},
  pages={79--89},
  year={2023}
}

@inproceedings{jiang2023gaussianshader,
  title={Gaussian{S}hader: 3{D} {G}aussian splatting with shading functions for reflective surfaces},
  author={Jiang, Yingwenqi and Tu, Jiadong and Liu, Yuan and Gao, Xifeng and Long, Xiaoxiao and Wang, Wenping and Ma, Yuexin},
  booktitle={CVPR},
  pages={5322--5332},
  year={2024}
}

@article{gao2023relightable,
  title={Relightable 3{D} {G}aussian: Real-time point cloud relighting with {BRDF} decomposition and ray tracing},
  author={Gao, Jian and Gu, Chun and Lin, Youtian and Zhu, Hao and Cao, Xun and Zhang, Li and Yao, Yao},
  journal={arXiv preprint arXiv:2311.16043},
  year={2023}
}

@inproceedings{zeng2023relighting,
  title={Relighting neural radiance fields with shadow and highlight hints},
  author={Zeng, Chong and Chen, Guojun and Dong, Yue and Peers, Pieter and Wu, Hongzhi and Tong, Xin},
  booktitle={ACM SIGGRAPH},
  pages={1--11},
  year={2023}
}

@article{kerbl20233d,
  title={3{D} {G}aussian splatting for real-time radiance field rendering},
  author={Kerbl, Bernhard and Kopanas, Georgios and Leimk{\"u}hler, Thomas and Drettakis, George},
  journal={TOG},
  volume={42},
  number={4},
  pages={1--14},
  year={2023},
  publisher={ACM}
}

@article{mildenhall2021nerf,
  title={Ne{RF}: Representing scenes as neural radiance fields for view synthesis},
  author={Mildenhall, Ben and Srinivasan, Pratul P and Tancik, Matthew and Barron, Jonathan T and Ramamoorthi, Ravi and Ng, Ren},
  journal={Communications of the ACM},
  volume={65},
  number={1},
  pages={99--106},
  year={2021},
  publisher={ACM New York, NY, USA}
}

@InProceedings{Srinivasan_2021_CVPR,
    author    = {Srinivasan, Pratul P. and Deng, Boyang and Zhang, Xiuming and Tancik, Matthew and Mildenhall, Ben and Barron, Jonathan T.},
    title     = {Ne{RV}: Neural Reflectance and Visibility Fields for Relighting and View Synthesis},
    booktitle = {CVPR},
    month     = {June},
    year      = {2021},
    pages     = {7495-7504}
}

@InProceedings{Zhang_2021_CVPR,
    author    = {Zhang, Kai and Luan, Fujun and Wang, Qianqian and Bala, Kavita and Snavely, Noah},
    title     = {Phy{SG}: Inverse Rendering With Spherical {G}aussians for Physics-Based Material Editing and Relighting},
    booktitle = {CVPR},
    month     = {June},
    year      = {2021},
    pages     = {5453-5462}
}

@InProceedings{Toschi_2023_CVPR,
    author    = {Toschi, Marco and De Matteo, Riccardo and Spezialetti, Riccardo and De Gregorio, Daniele and Di Stefano, Luigi and Salti, Samuele},
    title     = {ReLight {M}y {N}eRF: A Dataset for Novel View Synthesis and Relighting of Real World Objects},
    booktitle = {CVPR},
    month     = {June},
    year      = {2023},
    pages     = {20762-20772}
}

@InProceedings{Zheng_2023_CVPR,
    author    = {Zheng, Ruichen and Li, Peng and Wang, Haoqian and Yu, Tao},
    title     = {Learning Visibility Field for Detailed 3{D} Human Reconstruction and Relighting},
    booktitle = {CVPR},
    month     = {June},
    year      = {2023},
    pages     = {216-226}
}

@InProceedings{Wang_2023_ICCV,
    author    = {Wang, Dongqing and Zhang, Tong and S\"usstrunk, Sabine},
    title     = {{NEMTO}: Neural Environment Matting for Novel View and Relighting Synthesis of Transparent Objects},
    booktitle = {ICCV},
    month     = {October},
    year      = {2023},
    pages     = {317-327}
}

@inproceedings{sun2023sol,
  title={{SOL}-{NeRF}: Sunlight modeling for outdoor scene decomposition and relighting},
  author={Sun, Jia-Mu and Wu, Tong and Yang, Yong-Liang and Lai, Yu-Kun and Gao, Lin},
  booktitle={ACM SIGGRAPH Asia},
  pages={1--11},
  year={2023}
}

@InProceedings{Chang_2024_WACV,
    author    = {Chang, Yeonjin and Kim, Yearim and Seo, Seunghyeon and Yi, Jung and Kwak, Nojun},
    title     = {Fast Sun-Aligned Outdoor Scene Relighting Based on {T}enso{RF}},
    booktitle = {WACV},
    month     = {January},
    year      = {2024},
    pages     = {3626-3636}
}

@InProceedings{Yang_2023_CVPR,
    author    = {Yang, Siqi and Cui, Xuanning and Zhu, Yongjie and Tang, Jiajun and Li, Si and Yu, Zhaofei and Shi, Boxin},
    title     = {Complementary Intrinsics From Neural Radiance Fields and {CNN}s for Outdoor Scene Relighting},
    booktitle = {CVPR},
    month     = {June},
    year      = {2023},
    pages     = {16600-16609}
}

@inproceedings{blinn1977models,
  title={Models of light reflection for computer synthesized pictures},
  author={Blinn, James F},
  booktitle={ACM SIGGRAPH},
  pages={192--198},
  year={1977}
}

@article{zhang2021nerfactor,
  title={Ne{RF}actor: Neural factorization of shape and reflectance under an unknown illumination},
  author={Zhang, Xiuming and Srinivasan, Pratul P and Deng, Boyang and Debevec, Paul and Freeman, William T and Barron, Jonathan T},
  journal={TOG},
  volume={40},
  number={6},
  pages={1--18},
  year={2021},
  publisher={ACM New York, NY, USA}
}

@inproceedings{wu2023nerf,
  title={{DE}-{N}e{RF}: Decoupled neural radiance fields for view-consistent appearance editing and high-frequency environmental relighting},
  author={Wu, Tong and Sun, Jia-Mu and Lai, Yu-Kun and Gao, Lin},
  booktitle={ACM SIGGRAPH},
  pages={1--11},
  year={2023}
}

@inproceedings{sarkar2023litnerf,
  title={Lit{N}e{RF}: Intrinsic Radiance Decomposition for High-Quality View Synthesis and Relighting of Faces},
  author={Sarkar, Kripasindhu and B{\"u}hler, Marcel C and Li, Gengyan and Wang, Daoye and Vicini, Delio and Riviere, J{\'e}r{\'e}my and Zhang, Yinda and Orts-Escolano, Sergio and Gotardo, Paulo and Beeler, Thabo and others},
  booktitle={ACM SIGGRAPH Asia},
  pages={1--11},
  year={2023}
}

@inproceedings{xu2023renerf,
  title={Re{N}e{RF}: Relightable neural radiance fields with nearfield lighting},
  author={Xu, Yingyan and Zoss, Gaspard and Chandran, Prashanth and Gross, Markus and Bradley, Derek and Gotardo, Paulo},
  booktitle={ICCV},
  pages={22581--22591},
  year={2023}
}

@article{paszke2019pytorch,
  title={Pytorch: An imperative style, high-performance deep learning library},
  author={Paszke, Adam and Gross, Sam and Massa, Francisco and Lerer, Adam and Bradbury, James and Chanan, Gregory and Killeen, Trevor and Lin, Zeming and Gimelshein, Natalia and Antiga, Luca and others},
  journal={NeurIPS},
  volume={32},
  year={2019}
}

@article{kingma2014adam,
  title={Adam: A method for stochastic optimization},
  author={Kingma, Diederik P and Ba, Jimmy},
  journal={arXiv preprint arXiv:1412.6980},
  year={2014}
}

@inproceedings{zhang2018unreasonable,
  title={The unreasonable effectiveness of deep features as a perceptual metric},
  author={Zhang, Richard and Isola, Phillip and Efros, Alexei A and Shechtman, Eli and Wang, Oliver},
  booktitle={CVPR},
  pages={586--595},
  year={2018}
}

@article{wang2004image,
  title={Image quality assessment: from error visibility to structural similarity},
  author={Wang, Zhou and Bovik, Alan C and Sheikh, Hamid R and Simoncelli, Eero P},
  journal={IEEE transactions on image processing},
  volume={13},
  number={4},
  pages={600--612},
  year={2004},
  publisher={IEEE}
}

@incollection{phong1998illumination,
  title={Illumination for computer generated pictures},
  author={Phong, Bui Tuong},
  booktitle={Seminal graphics: pioneering efforts that shaped the field},
  pages={95--101},
  year={1998}
}

@article{yariv2021volume,
  title={Volume rendering of neural implicit surfaces},
  author={Yariv, Lior and Gu, Jiatao and Kasten, Yoni and Lipman, Yaron},
  journal={NeurIPS},
  volume={34},
  pages={4805--4815},
  year={2021}
}

@article{yu2022monosdf,
  title={Mono{SDF}: Exploring monocular geometric cues for neural implicit surface reconstruction},
  author={Yu, Zehao and Peng, Songyou and Niemeyer, Michael and Sattler, Torsten and Geiger, Andreas},
  journal={NeurIPS},
  volume={35},
  pages={25018--25032},
  year={2022}
}

@article{zhang2020nerf++,
  title={Ne{RF}++: Analyzing and improving neural radiance fields},
  author={Zhang, Kai and Riegler, Gernot and Snavely, Noah and Koltun, Vladlen},
  journal={arXiv preprint arXiv:2010.07492},
  year={2020}
}

@article{liu2020neural,
  title={Neural sparse voxel fields},
  author={Liu, Lingjie and Gu, Jiatao and Zaw Lin, Kyaw and Chua, Tat-Seng and Theobalt, Christian},
  journal={NeurIPS},
  volume={33},
  pages={15651--15663},
  year={2020}
}

@inproceedings{luiten2023dynamic,
  title={Dynamic 3{D} {G}aussians: Tracking by persistent dynamic view synthesis},
  author={Luiten, Jonathon and Kopanas, Georgios and Leibe, Bastian and Ramanan, Deva},
  booktitle={3DV},
  pages={800--809},
  year={2024},
  organization={IEEE}
}

@inproceedings{wu20234d,
  title={4{D} {G}aussian splatting for real-time dynamic scene rendering},
  author={Wu, Guanjun and Yi, Taoran and Fang, Jiemin and Xie, Lingxi and Zhang, Xiaopeng and Wei, Wei and Liu, Wenyu and Tian, Qi and Wang, Xinggang},
  booktitle={CVPR},
  pages={20310--20320},
  year={2024}
}

@inproceedings{yang2023deformable,
  title={Deformable 3{D} {G}aussians for high-fidelity monocular dynamic scene reconstruction},
  author={Yang, Ziyi and Gao, Xinyu and Zhou, Wen and Jiao, Shaohui and Zhang, Yuqing and Jin, Xiaogang},
  booktitle={CVPR},
  pages={20331--20341},
  year={2024}
}

@article{chen2020closer,
  title={A closer look at the training strategy for modern meta-learning},
  author={Chen, Jiaxin and Wu, Xiao-Ming and Li, Yanke and Li, Qimai and Zhan, Li-Ming and Chung, Fu-lai},
  journal={NeurIPS},
  volume={33},
  pages={396--406},
  year={2020}
}

@inproceedings{zhang2022iron,
  title={{IRON}: Inverse rendering by optimizing neural sdfs and materials from photometric images},
  author={Zhang, Kai and Luan, Fujun and Li, Zhengqi and Snavely, Noah},
  booktitle={CVPR},
  pages={5565--5574},
  year={2022}
}

@inproceedings{cheng2023wildlight,
  title={Wild{L}ight: In-the-wild inverse rendering with a flashlight},
  author={Cheng, Ziang and Li, Junxuan and Li, Hongdong},
  booktitle={CVPR},
  pages={4305--4314},
  year={2023}
}

@inproceedings{liang2024gs,
  title={G{S}-{IR}: 3{D} {G}aussian splatting for inverse rendering},
  author={Liang, Zhihao and Zhang, Qi and Feng, Ying and Shan, Ying and Jia, Kui},
  booktitle={CVPR},
  pages={21644--21653},
  year={2024}
}

@article{fan2024rng,
  title={{RNG}: Relightable Neural {G}aussians},
  author={Fan, Jiahui and Luan, Fujun and Yang, Jian and Ha{\v{s}}an, Milo{\v{s}} and Wang, Beibei},
  journal={arXiv preprint arXiv:2409.19702},
  year={2024}
}

@inproceedings {bi2024gs,
    title      = {G{S}\textsuperscript{3}: Efficient Relighting with Triple Gaussian Splatting},
    author     = {Zoubin Bi and Yixin Zeng and Chong Zeng and Fan Pei and Xiang Feng and Kun Zhou and Hongzhi Wu},
    booktitle  = {ACM SIGGRAPH Asia},
    year       = {2024}
}

@article{zhu2024gs,
  title={Gs-ror: 3d gaussian splatting for reflective object relighting via sdf priors},
  author={Zhu, Zuo-Liang and Wang, Beibei and Yang, Jian},
  journal={arXiv preprint arXiv:2406.18544},
  year={2024}
}

@inproceedings{kuang2024olat,
  title={OLAT Gaussians for Generic Relightable Appearance Acquisition},
  author={Kuang, Zhiyi and Yang, Yanchao and Dong, Siyan and Ma, Jiayue and Fu, Hongbo and Zheng, Youyi},
  booktitle={SIGGRAPH Asia 2024 Conference Papers},
  pages={1--11},
  year={2024}
}
}

% References follow the acknowledgments in the camera-ready paper. Use unnumbered first-level heading for the references. Any choice of citation style is acceptable as long as you are consistent. It is permissible to reduce the font size to \verb+small+ (9 point)
% when listing the references.
% Note that the Reference section does not count towards the page limit.
% \medskip

% {
% \small

% [1] Alexander, J.A.\ \& Mozer, M.C.\ (1995) Template-based algorithms for
% connectionist rule extraction. In G.\ Tesauro, D.S.\ Touretzky and T.K.\ Leen
% (eds.), {\it Advances in Neural Information Processing Systems 7},
% pp.\ 609--616. Cambridge, MA: MIT Press.

% [2] Bower, J.M.\ \& Beeman, D.\ (1995) {\it The Book of GENESIS: Exploring
%   Realistic Neural Models with the GEneral NEural SImulation System.}  New York:
% TELOS/Springer--Verlag.

% [3] Hasselmo, M.E., Schnell, E.\ \& Barkai, E.\ (1995) Dynamics of learning and
% recall at excitatory recurrent synapses and cholinergic modulation in rat
% hippocampal region CA3. {\it Journal of Neuroscience} {\bf 15}(7):5249-5262.
% }

%%%%%%%%%%%%%%%%%%%%%%%%%%%%%%%%%%%%%%%%%%%%%%%%%%%%%%%%%%%%

\clearpage
\setcounter{page}{1}
\appendix

% \section{Technical Appendices and Supplementary Material}

% {
%     \centering
%     \Large
%     \textbf{MetaGS: A Meta-Learned Gaussian-Phong Model for Out-of-Distribution 3D Scene Relighting}\\
%     \vspace{0.5em}Supplementary Material \\
%     \vspace{1.0em}
% }

\appendix

\section{Notations}
The symbols and notations are illustrated in \tabref{tab:notations}. 

\begin{table}[h]
\centering
% \vspace{-5pt}
\caption{Notations in the manuscript. $^\dagger$Per-point variables unless being marked as \textit{global}.}
\small
\vspace{-3pt}
\setlength{\tabcolsep}{20pt}
\begin{tabular}{lcc}
\toprule
\textbf{Location} & \textbf{Notations} & \textbf{Meaning} \\ \midrule
\multirow{8}{*}{Phong model$^\dagger$}
    & $I$ & learnable light intensity (\textit{global}) \\
    & $T_i^{light}$ & received light intensity \\ 
    & $k_d, k_s$ & diffuse/specular coefficient \\ 
    & $I_d, I_s$ & diffuse/specular intensity \\
    & $r$ & point-to-light distance \\ 
    & $\mathbf n$ & normalized normal vector \\ 
    & $\mathbf v$ & point-to-camera normalized vector \\
    & $\mathbf l$ & point-to-light normalized vector \\ \midrule
\multirow{2}{*}{Meta-learning} 
    % & $m$ & task number per iteration \\
    & $\theta$ & Gaussian attributes \\
    % & $\phi$ & shadow coefficient \\
    % & $\theta_i', \phi_i'$ & i-th task-specific predictor 
    & $\theta_i'$ & i-th task-specific predictor 
    \\ \bottomrule
\end{tabular}
% \end{small}
% }
\label{tab:notations}
\vspace{-5pt}
\end{table}

\section{Datasets}

% We perform three OLAT relighting experiments: out-of-distribution relighting, in-distribution novel view relighting, and camera-light-colocated relighting.

\begin{figure}[h]
    \centering
    \includegraphics[width=\linewidth]{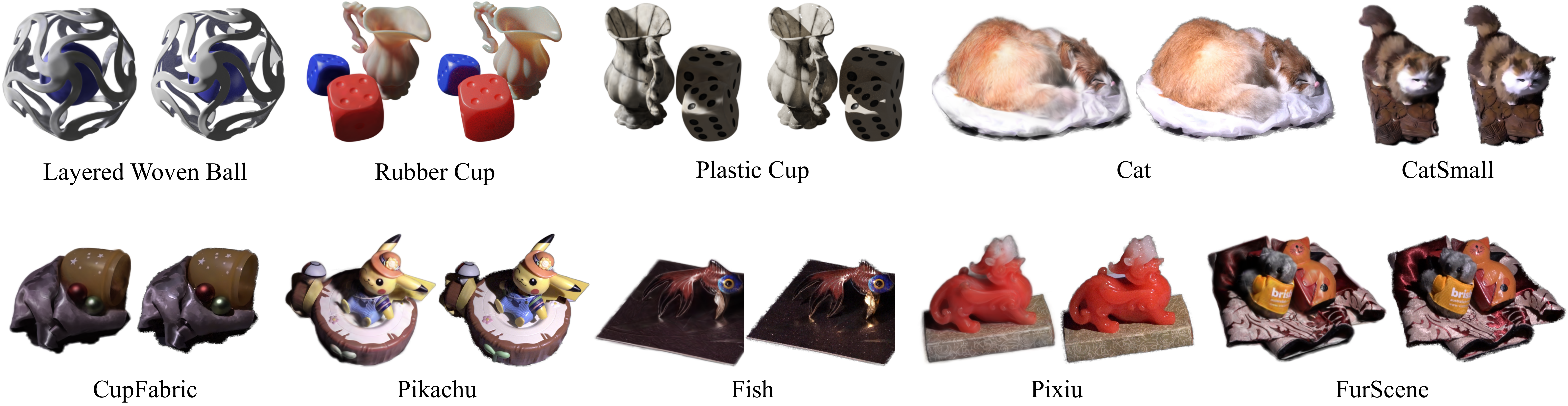}
    \caption{Synthetic and real-world scenes used in the experiment with ground-truth images (right) and rendering results of \model{} (left).}
    \label{sections/fig:data}
    \vspace{-5pt}
\end{figure}

We test on three synthetic scenes and seven real-world scenes from NRHints~\cite{zeng2023relighting}, as shown in \figref{sections/fig:data}.
Each image is rendered (or captured) with a unique camera pose and point-light position.
For each synthetic scene, we generate a total of $600$ OLAT images using Blender, with $500$ for training and $100$ for testing, following either out-of-distribution or colocated data patterns to align with their settings.
Specifically, the camera and light source are independently sampled on the upper hemisphere with centered around the scene, with elevation angles ranging from $10^{\circ}$ to $80^{\circ}$, and radial distances uniformly sampled between $4.0$× and $5.0$× the object bounding size. This setup ensures a wide diversity of incident angles while maintaining plausible visibility. 
In the OOD configuration, training and testing light directions are drawn from non-overlapping hemispheres (azimuth $\phi\in[0,\pi]$ for training; $\phi\in[\pi,2\pi]$ for testing) to induce generalization difficulty.
In the colocated setting, camera and light positions are identical, emulating a mobile-light capture scenario.
All images are rendered at a resolution of $512\times512$ using Cycles path tracing in Blender, with $256$ samples per pixel.
For real captured scenes in the out-of-distribution experiment, we manually divide the data into training and testing sets, each with lights positioned on opposite sides of the upper hemisphere.
% The \textit{Pikachu} scene features glossy highlights and significant self-occlusion. 
% The \textit{Cat} scene presents complex geometry on the cat’s fur. 
% The \textit{CupFabric} scene showcases translucent materials on the cup, specular reflections of the balls, and anisotropic reflections on the fabric. 
% The \textit{Fish} scene involves intricate secondary reflections between objects and the ground.

% Different settings are explained in Table \tabref{tab:setup}.

\begin{table}[t]
\centering
\caption{Full comparison of the OOD relighting comparison on NRHints synthetic datasets.}
\small
\vspace{-5pt}
% \scalebox{0.9}{
% \begin{adjustbox}{center}
% \resizebox{\textwidth}{!}{
\begin{tabular}{lccccccccc}
\toprule
\multirow{2.5}{*}{Method} & \multicolumn{3}{c}{Layered Woven Ball} & \multicolumn{3}{c}{Plastic Cup} & \multicolumn{3}{c}{Rubber Cup}  \\ %\midrule
 \cmidrule(lr){2-4} \cmidrule(lr){5-7}  \cmidrule(lr){8-10} 
 & PSNR & SSIM & LPIPS & PSNR & SSIM & LPIPS & PSNR & SSIM & LPIPS \\ \midrule
NRHints~\cite{zeng2023relighting} &   17.25&0.8755&0.0872   &   23.92&0.9485& 0.0470  &   27.44&0.9465&0.0691    \\ 
WildLight~\cite{cheng2023wildlight} &    21.73&0.9217&0.0568    &    20.95&0.9249&0.0629   &   24.02&0.9221&0.0789    \\
GS$^3$~\cite{bi2024gs} &   18.84&0.9063&0.0589   &   20.30&0.9141&0.0628   &   24.37&0.9349&0.0715        \\
Ours &  \textbf{26.76}&\textbf{0.9561}&\textbf{0.0423}  &  
\textbf{27.54}&\textbf{0.9580}&\textbf{0.0405}  &
\textbf{27.95}&\textbf{0.9497}&\textbf{0.0688}
\\ \bottomrule
\end{tabular}
% }
% \end{adjustbox}
% }
\label{tab:ood_synthetic_all}
\vspace{-5pt}
\end{table}

\begin{table}[t]
\centering
\caption{Full comparison of OOD relighting results on NRHints real-world datasets.}
% \small
\vspace{-5pt}
\resizebox{\textwidth}{!}{
\begin{tabular}{lcccccccccccc}
\toprule
\multirow{2.5}{*}{Method} & \multicolumn{3}{c}{Cat} & \multicolumn{3}{c}{Catsmall} & \multicolumn{3}{c}{CupFabric}  & \multicolumn{3}{c}{Fish} \\ % Upper Part
\cmidrule(lr){2-4} \cmidrule(lr){5-7}  \cmidrule(lr){8-10} \cmidrule(lr){11-13} 
 & PSNR & SSIM & LPIPS & PSNR & SSIM & LPIPS & PSNR & SSIM & LPIPS & PSNR & SSIM & LPIPS \\ \midrule
NRHints~\cite{zeng2023relighting}   & 18.04 & 0.7821 & 0.2405 & 24.63 & 0.9289 & 0.1094 & 24.65 & 0.9377 & 0.0965 & 22.57 & 0.8658 & 0.1396 \\ 
WildLight~\cite{cheng2023wildlight} & 18.65 & 0.7293 & 0.2685 & 22.53 & 0.8820 & 0.1448 & 24.03 & 0.9246 & 0.0969 & 21.47 & 0.8419 & 0.1458 \\ 
GS$^3$~\cite{bi2024gs}    & 17.66 & 0.6919 & 0.2431 & 23.34 & 0.9182 & 0.1097 & 25.04 & 0.9408 & 0.0954 & 21.12 & 0.8264 & 0.1401 \\ 
Ours     
& \textbf{26.45} & \textbf{0.8514} & \textbf{0.1759}
& \textbf{26.44} & \textbf{0.9505} & \textbf{0.0980}
& \textbf{27.29} & \textbf{0.9490} & \textbf{0.0918}
& \textbf{24.68} & \textbf{0.8706} & \textbf{0.1386} \\ 
% \bottomrule\\
% \toprule
\midrule
\multirow{2.5}{*}{Method} & \multicolumn{3}{c}{FurScene} & \multicolumn{3}{c}{Pikachu} & \multicolumn{3}{c}{Pixiu} & \multicolumn{3}{c}{Average} \\ % Lower Part
\cmidrule(lr){2-4} \cmidrule(lr){5-7} \cmidrule(lr){8-10} \cmidrule(lr){11-13} 
 & PSNR & SSIM & LPIPS & PSNR & SSIM & LPIPS & PSNR & SSIM & LPIPS & PSNR & SSIM & LPIPS \\ \midrule
NRHints~\cite{zeng2023relighting}   & 21.55 & 0.8568 & 0.1435 & 24.00 & 0.9279 & 0.0958 & 23.03 & 0.8970 & 0.1203 & 22.64&0.8853&0.1351 \\ 
WildLight~\cite{cheng2023wildlight} & 20.33 & 0.8304 & 0.1635 & 19.09 & 0.8941 & 0.1129 & 20.22 & 0.8330 & 0.1621 & 20.90&0.8479&0.1564 \\ 
GS$^3$~\cite{bi2024gs}    & 17.34 & 0.7684 & 0.1669 & 24.11 & 0.9203 & 0.0927 & 19.63 & 0.8391 & 0.1329 & 21.18&0.8436&0.1387 \\ 
Ours      
& \textbf{24.82} & \textbf{0.8850} & \textbf{0.1297}
& \textbf{25.54} & \textbf{0.9350} & \textbf{0.0928}
& \textbf{25.65}& \textbf{0.9194} & \textbf{0.1105} 
& \textbf{25.83}&\textbf{0.9087}&\textbf{0.1196} \\ 
\bottomrule
\end{tabular}
}
\label{tab:ood_real_world_all}
\vspace{-10pt}
\end{table}

\section{More Experimental Analyses}

\subsection{Full Comparisons for OOD Relighting}

We present the OOD relighting results with all three metrics in \tabref{tab:ood_synthetic_all} and \tabref{tab:ood_real_world_all}.

\subsection{Clarification on Meta-learning}

% % insight: 在machine learning中，需要将数据集分成trainset和testset，但是，在trainset上的训练过程可能导致过拟合，进而引起在testset上效果较差。
% % Meta-learning是一种面向testset效果的训练方法，核心是在训练的过程中模拟测试环境，从而提升模型在测试集上的效果。
% % 回到我们的光照模型任务。光照模型也会倾向于过拟合到某一种光照下的场景表达。因此，我们creatively将每种光照下的场景学习视为不同的任务，通过meta-learning来避免过拟合、加速收敛、提升模型在测试集上的效果。

% % By updating the model with loss functions from distinct tasks, the model acquires a better overall capability and can be better fine-tuned for a wider range of unseen tasks.

% In the field of machine learning, overfitting is a common issue. Meta-learning addresses this problem by \textbf{simulating test conditions during training to enhance testset performance}. It organizes training data into tasks, each with its own loss function. During training, the model continuously switches between different tasks, learning to generalize better to new, unseen data. In 2020, ~\cite{chen2020closer} mathematically demonstrated that meta-learning reduces the distribution discrepancy between the training set and testing set, showing that \textit{with a sufficient number of training tasks, the generalization gap can converge to zero, regardless of the number of samples per task}.

% For illumination modelling, the model tends to overfit to specific lighting conditions. To address this, we creatively treat the learning under each lighting condition as a separate task and apply meta-learning to avoid overfitting, accelerate geometry-texture decomposition, and improve the model's testing performance.

In our OLAT method, meta-learning helps mitigate overfitting to specific illuminations by explicitly simulating test conditions during training.
%
% The core benefit of meta-learning lies in its ability to mitigate overfitting through explicit simulation of test conditions during training. 
%
% In our tasks, conventional OLAT training tends to induce high-frequency geometric overfitting to specific illumination patterns (e.g., misattributing local brightness variations to wrong geometric details). 
%
%
Consider a renderer $R(\theta, l_i)$, where $\theta$ represents light-independent parameters (material/geometry), and $l_i$ denotes the light position. 
% Formally, consider our rendering model as $R_\theta(A, l)$, where $A$ denotes appearance parameters (material/geometry) and $l$ represents light position. 
%
% Conventional OLAT optimization minimizes $\mathcal{L}=$ $\|R(A, l)-I\|^2$ directly, allowing geometry parameters to encode illumination-specific artifacts rather than true material properties.
%
Through bilevel gradients, meta-learning assesses how well the learned $\theta'$ under $L=\|R(\theta, l_i)-I_i\|^2$ adapts to minimizing $\|R(\theta', l_j)-I_j\|^2$ with the simulated test condition $l_j$, thereby enforcing cross-illumination consistency in geometry reconstruction. 
This new loss function encourages the model to learn to generalize to unseen illumination rather than to overfit to a specific training sample as direct RGB loss does.
By forcing geometry parameters $\theta$ to remain consistent across $\{l_i'\}$, we prevent illumination artifacts from being baked into textures, thereby achieving more accurate highlights and smoother surfaces.
As shown in \tabref{tab:larger_batch_size}, increasing batch size (by task number) improves PSNR by $2+$, while meta-learning achieves an $8+$ improvement, confirming that the impact of meta-learning is not due to a larger batch size. % ($m=5$).

\begin{table}[h]
\vspace{-5pt}
\caption{A comparison of meta-learning \textit{vs.} Increased batch size.}
\begin{center}
\small
\vspace{-5pt}
\setlength{\tabcolsep}{20pt}
    \begin{tabular}{lccc}
        \toprule
         & 
        Meta-learn &
        w/o Meta-Learn  &
        Larger batch ($\times m$)
        \\
        \midrule
        PSNR & 27.42 & 19.14 & 20.94 \\
        \bottomrule
    \end{tabular}
\end{center}
\label{tab:larger_batch_size}
\vspace{-10pt}
\end{table}

\subsection{Multi-light Relighting}

\begin{figure}[!h]
\vspace{-5pt}
  \centering
  \includegraphics[width=0.75\linewidth]{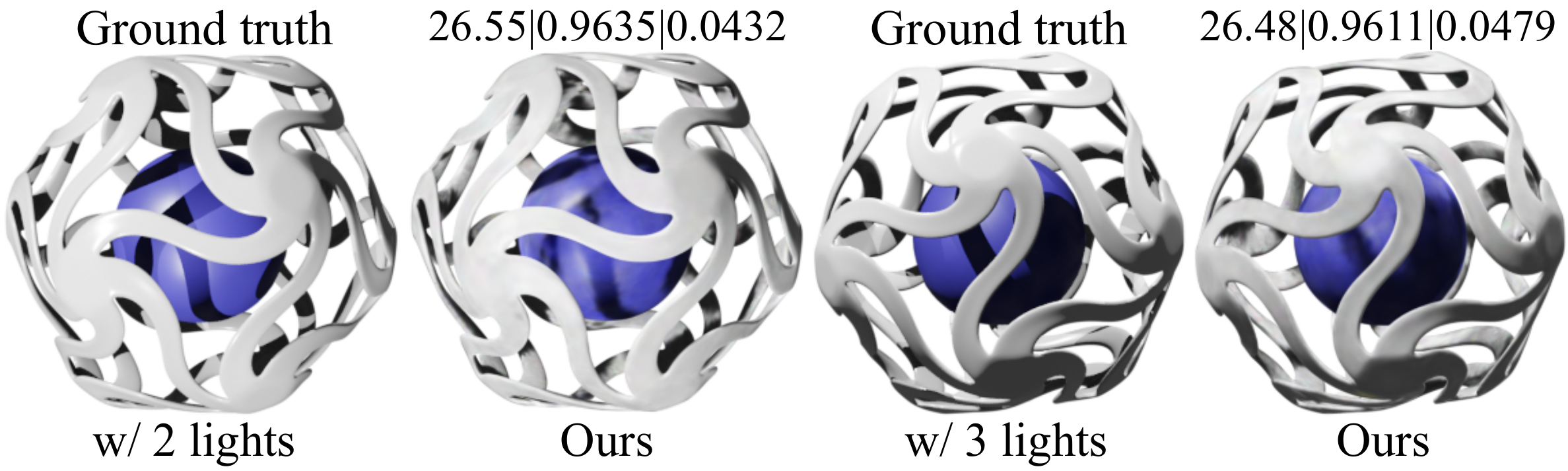}
  \vspace{-5pt}
   \caption{Results with multiple lights (PSNR/SSIM/LPIPS).}
   \label{fig:multilight}
   \vspace{-10pt}
\end{figure}

To validate \model{}'s capability in handling complex lighting conditions, we conduct experiments with test sets containing $2$--$3$ light sources. As shown in Figure \ref{fig:multilight}, our method demonstrates robust generalization beyond single-light scenarios.

\subsection{Camera-Light-Colocated Relighting}

\begin{figure}[t]
    \centering
    \includegraphics[width=0.8\linewidth]{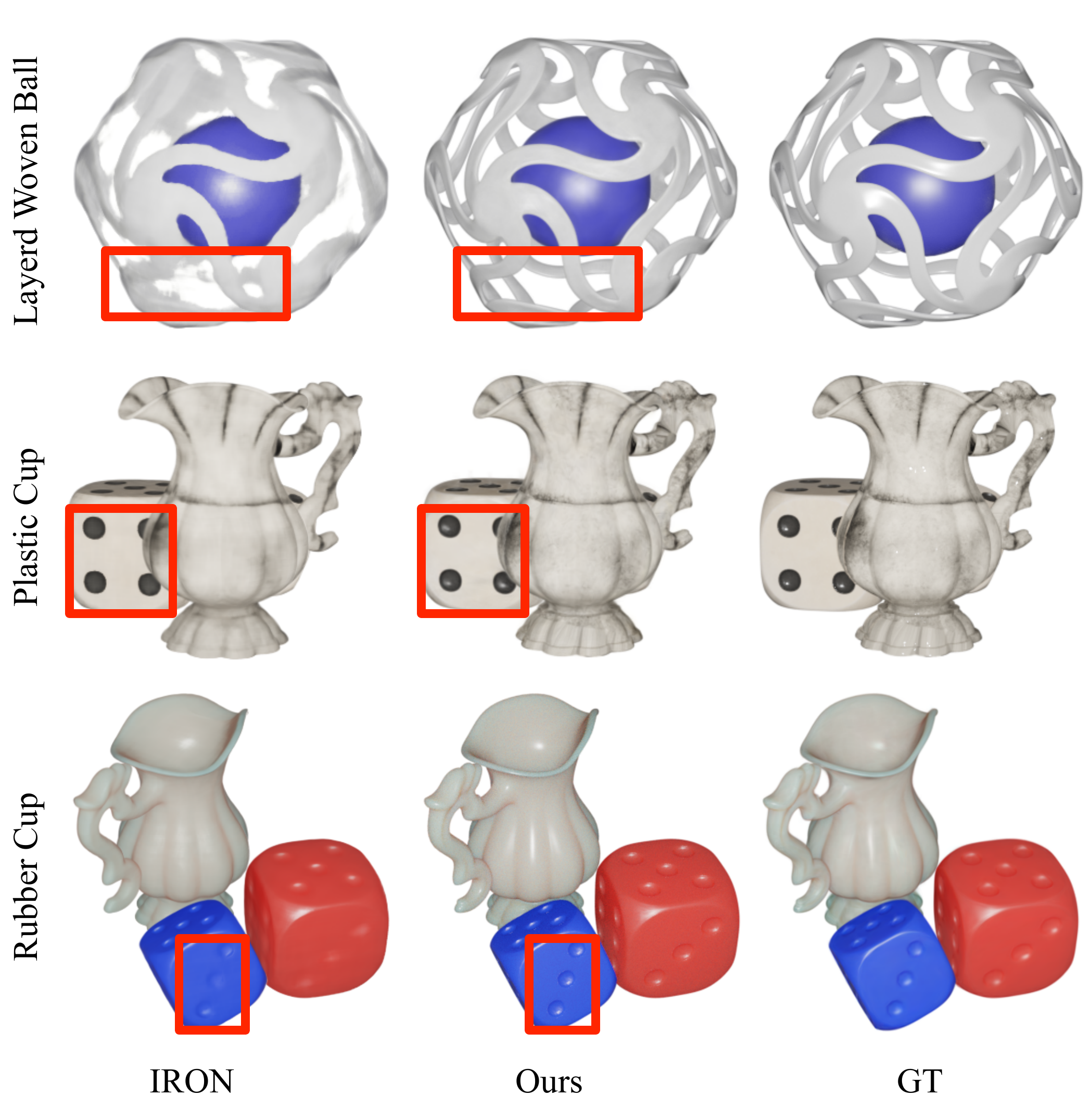}
    \vspace{-10pt}
    \caption{Novel view synthesis results under the camera-light-colocated relighting setup. Compared to IRON, our model more accurately infers object geometry in the Layered Woven Ball scene and produces fewer artifacts in the Plastic Cup scene.}
    \label{sections/fig:iron}
    \vspace{-5pt}
\end{figure}

We present the colocated relighting results with all three metrics in Table \ref{tab:syn_iron} and the qualitative results in \figref{sections/fig:iron}.
%
% Note that IRON fails to accurately infer the geometry of both the Complex Ball and Rubber Cup scenes during its initial volume learning stage. This leads to a model crash in the subsequent surface learning stage. Therefore, we use the results from the first stage for our calculations.
%
As shown in the figure, IRON fails to accurately infer the geometry of the Layered Woven Ball scene, resulting in blurry results. In the Plastic Cup and the Rubber Cup scenes, while IRON can provide reasonable geometry reconstruction, it shows more artifacts in the appearance modeling, such as the dent of the dice. On the other hand, \model{} can provide more accurate illumination details.

\begin{table}[t]
\caption{Novel view synthesis results in the \textbf{camera-light-colocated} relighting setup.}
\begin{center}
\vspace{-5pt}
\small
% \resizebox{\textwidth}{!}{
\setlength\tabcolsep{7pt}
    \begin{tabular}{lccccccccc}
        \toprule
        \multirow{2.5}{*}{Method} & \multicolumn{3}{c}{Layered Woven Ball} & \multicolumn{3}{c}{Plastic Cup} & \multicolumn{3}{c}{Rubber Cup}\\
        \cmidrule(lr){2-4} \cmidrule(lr){5-7} \cmidrule(lr){8-10} 
        % & PSNR$^\uparrow$ & SSIM$^\uparrow$ & LPIPS$^\downarrow$ & PSNR$^\uparrow$ & SSIM$^\uparrow$ & LPIPS$^\downarrow$S& PSNR$^\uparrow$ & SSIM$^\uparrow$ & LPIPS$^\downarrow$  \\
        & PSNR & SSIM & LPIPS & PSNR & SSIM & LPIPS& PSNR & SSIM & LPIPS\\
        \midrule
        IRON~\cite{zhang2022iron}  &  26.99&0.9231&0.0896  &    34.43&0.9808&0.0296  &  36.22&0.9677&0.0476      \\
        % NRHints~\cite{}  &  27.88&0.9464&0.0485  &    \textbf{39.85}&\textbf{0.9009}&0.0185  &  \textbf{41.33}&\textbf{0.9733}&0.0386      \\
        Ours   &   \textbf{38.72}&\textbf{0.9904}&\textbf{0.0124}   &   \textbf{36.90}&\textbf{0.9893}&\textbf{0.0145}  &   \textbf{38.89}&\textbf{0.9712}&\textbf{0.0385} \\
        \bottomrule
    \end{tabular}
% }
\end{center}
\label{tab:syn_iron}
\vspace{-10pt}
\end{table}

\subsection{In-distribution Novel View Relighting}

This experiment follows the setup in NRHints~\cite{zeng2023relighting}, where both training and testing lights are randomly located in the upper hemisphere surrounding the scene). 
Each view is illuminated by a single point light placed at a unique position, though all light positions are distributed within the same region.
During the testing phase, we evaluate on novel viewpoints under \textit{new, in-distribution} light positions. 
Similar to the OOD setting, we generate $600$ OLAT images per synthetic scene using Blender ($500$ for training and $100$ for testing).
For real-world scenes, we use only $600$ training images--—a significantly smaller subset compared to the full NRHints dataset.

We present the quantitative results for free-viewpoint relighting with in-distribution point light positions in \tabref{tab:nvs_synthetic}. % and \tabref{tab:nvsrealworld}. 
Notably, \model{} outperforms other near real-time methods, including Relightable 3DGS and GaussianShader, by large margins, and shows an advantage over GS$^3$, a Gaussian-based OLAT method.
Our method underperforms NRHints on certain ``not-so-difficult'' datasets such as \textit{PlasticCup} and \textit{RubberCup}. 
This is partly because the implicit rendering model in NRHints provides better color fitting, resulting in superior in-distribution performance. While the Phong-based rendering model used in our work facilitates geometry-shading decoupling and is particularly effective for objects with incoherent geometries, such as the \textit{Woven Ball}.
However, does NRHints truly learn the intrinsic reflection mechanisms from the observed images? Likely not, as evidenced in the previous OOD results.

%GS^3: didn't use camera position and light position refinement for a fair comparison. adventage over other 3dgs-based methods including olat methods gs$^3$.

\begin{table}[t]
\caption{Novel view synthesis results in the in-distribution relighting setup. We present the best and second-best results in bold and underlined format, respectively.
}
\vspace{-10pt}
\setlength\tabcolsep{3.5pt}
% \small
\begin{center}
\resizebox{\textwidth}{!}{
    \begin{tabular}{lcccccccccccc}
        \toprule
        \multirow{2.5}{*}{Method} & \multicolumn{3}{c}{Layered Woven Ball} & \multicolumn{3}{c}{Plastic Cup} & \multicolumn{3}{c}{Rubber Cup} & \multicolumn{3}{c}{Average} \\
        \cmidrule(lr){2-4} \cmidrule(lr){5-7} \cmidrule(lr){8-10} \cmidrule(lr){11-13}
        & PSNR$^\uparrow$ & SSIM$^\uparrow$ & LPIPS$^\downarrow$ & PSNR$^\uparrow$ & SSIM$^\uparrow$ & LPIPS$^\downarrow$ & PSNR$^\uparrow$ & SSIM$^\uparrow$ & LPIPS$^\downarrow$ & PSNR$^\uparrow$ & SSIM$^\uparrow$ & LPIPS$^\downarrow$ \\
        \midrule
        3DGS~\cite{kerbl20233d}                         &  19.55&0.9130&0.0580  &   21.13&0.9417&0.0470  &  23.22&0.9248&0.0796   &   21.30&0.9265&0.0615    \\
        Relightable 3DGS~\cite{gao2023relightable}      &  19.62&0.9088&0.0610  &   21.07&0.9390&0.0502  &  23.20&0.9224&0.0838   &   21.30&0.9234&0.0650    \\
        GaussianShader~\cite{jiang2023gaussianshader}   &  19.62&0.9137&0.0586  &   20.92&0.9394&0.0500  &  23.12&0.9241&0.0833   &   21.22&0.9257&0.0640    \\
        NRHints~\cite{zeng2023relighting}               &  18.99&0.9087&0.0731  &   \textbf{35.81}&\textbf{0.9847}&\textbf{0.0281}  &  \textbf{36.64}&\textbf{0.9637}&\textbf{0.0481}  &  \underline{30.48}&\underline{0.9524}&\underline{0.0498}     \\
        WildLight~\cite{cheng2023wildlight}             &  22.71&0.9284&0.0499  &  23.28&0.9068&0.1018  &  24.65&0.9271&0.0737  &  23.55&0.9208&0.0751  \\
        GS$^3$~\cite{bi2024gs}                          &  \underline{28.20}&\underline{0.9659}&\underline{0.0396} &29.53&0.9755&0.0317 &29.34&0.9520&0.0618 &29.02&0.9645&0.0444 \\ 
        Ours
        &   \textbf{30.88}&\textbf{0.9690}&\textbf{0.0281}
        &   \underline{32.51}&\underline{0.9814}&\underline{0.0295}
        &   \underline{31.75}&\underline{0.9577}&\underline{0.0570}
        &   \textbf{31.71}&\textbf{0.9694}&\textbf{0.0382}  
        \\
        \bottomrule
    \end{tabular}
}
\label{tab:nvs_synthetic}
\end{center}
\vspace{-5pt}
\end{table}

\section{Preliminaries}

\subsection{3D Gaussian Splatting}

Our method builds upon 3D Gaussian Splatting (3DGS)~\cite{kerbl20233d}, which employs explicit Gaussian points for 3D scene representation. 
Each Gaussian's geometry is characterized by an opacity $o\in[0,1]$, a center point $\mu\in\mathbb{R}^{3\times1}$ and a covariance matrix $\Sigma\in\mathbb{R}^{3\times3}$:
\begin{equation}
    G(\mathbf X)=e^{-\frac{1}{2}(\mathbf X-\mu)^T \Sigma^{-1}(\mathbf X-\mu)},
\end{equation}
where the covariance matrix $\Sigma$ can be decomposed into a scaling matrix $ S\in\mathbb{R}^{3\times1}$ and a quaternion-parameterized rotation matrix $ R\in\mathbb{R}^{3\times3}$ for differentiable optimization:
\begin{equation}
    \Sigma=R S S^T R^T.
\end{equation}

To render an image from a given viewpoint, 3DGS uses the splatting technique to cast Gaussians within the view frustum onto the camera plane.
The projected 2D covariance matrix $\Sigma^{\prime}$ in camera coordinates can be computed as $\Sigma^{\prime}=J W \Sigma W^T J^T$, where $ W$ and $J$ denote the viewing transformation and the Jacobian of the affine approximation of the projective transformation, respectively.
The color of each pixel $\mathbf p$ is calculated by alpha-blending $N$ ordered Gaussians $\{G_i|i=1,\cdots,N\}$ overlapping $\mathbf p$ as:
\begin{equation}
    \mathcal{C}=\sum_{i \in N} T_i \alpha_i \boldsymbol{c}_i \text{  with  }T_i=\displaystyle\prod_{j=1}^{i-1}\left(1-\alpha_j\right),
\end{equation}
where $\alpha_i$ denotes the alpha value calculated by multiplying the opacity $o$ with the 2D Gaussian contribution value derived from $\Sigma^{\prime}$, and $T_i$ denotes the accumulated transmittance along the ray.

Specifically, each Gaussian's color is defined by spherical harmonics (SH) coefficients $\mathcal{C} \in \mathbb{R}^k$ (where $k$ represents the degrees of freedom). 
While high-order SH coefficients can contribute to the overall lighting of a scene under various view directions, they are not specifically designed to handle view-dependent color directly, such as specular reflections and shadows.

\subsection{Blinn-Phong Reflection Model}

The Phong model is widely adopted in computer graphics for simulating the interaction of light with object surfaces~\cite{phong1998illumination}.
% , offering a balance between computational efficiency and visual realism.
%
In this model, the reflected light transport on a surface has three components: ambient ($L_a$), diffuse ($L_d$), and specular reflections ($L_s$).
The ambient reflection component represents the constant illumination present in the environment, simulating how light scatters and reflects off other surfaces to establish a baseline brightness level.
The diffuse reflection component, governed by Lambertian diffusion law, illustrates the scattering of light in multiple directions upon striking a rough surface, leading to a matte appearance.
This component is calculated based on the angle between the light direction ($\mathbf{l}$) and the surface normal ($\mathbf{n}$), ensuring that surfaces facing the light source appear brighter.
The specular term is computed based on the angle between the light direction vector and the bisector ($\mathbf{h}$) between the view direction ($\mathbf{v}$) and the light direction ($\mathbf{l}$).
The Phong model also incorporates a shininess exponent that governs the spread of the specular highlight, allowing the model to represent the surface's glossiness. 
We select the Blinn-Phong model~\cite{blinn1977models} to provide the illumination decomposition priors to 3DGS for its simple yet strong physical prior and computation efficiency.
With little addition to the Gaussian's attribute, our model produces realistic lighting effects.

\section{Hyperparameters}
\label{sec:hyper}

\tabref{tab:hparams} presents the hyperparameters of \model{}.

\begin{table}[t]
\centering
\caption{Hyperparameters of \model{}.} 
\vspace{-3pt}
\setlength{\tabcolsep}{20pt}
\small
% \begin{adjustbox}{center}
\begin{tabular}{lcc}
\toprule
\textbf{Name} & \textbf{Notation} & \textbf{Value}\\
\midrule

learning rate & $\beta$ &  same as 3DGS~\cite{kerbl20233d} \\
meta-train inner learning rate & $\alpha$ & 0.01 \\
meta-train task number & $m$ & $5$ \\
RGB D-SSIM loss scale & $\lambda$ & 0.2 \\
sparse loss scale & $\lambda_{\text{sparse}}$  & 0.001 \\
normal prediction loss scale & $\lambda_{\text {n}}$ & 0.2 \\
normal residual loss scale & $\lambda_{\text {res}}$ & 0.001 \\
flatten loss scale & $\lambda_{\text {s}}$ & 0.001 \\
% smooth loss scale & $\lambda_{\text{smooth}}$ & 0.1 \\
% diffuse prior loss scale & $\lambda_{\text {diffuse}}$ & 0.005 \\
\bottomrule
\end{tabular}
% \end{adjustbox}
% \end{small}
% }
\label{tab:hparams}
\end{table}

\end{document}